%% file: neurips_2024.tex
\documentclass{article}

\usepackage[square,numbers]{natbib}
\usepackage[final]{neurips_2024}
\usepackage{graphicx}
\usepackage{amsmath}
\usepackage{amsthm}
\usepackage{algorithm}
\usepackage[linesnumbered, algo2e]{algorithm2e}

\newtheorem{proposition}{Proposition}

\usepackage[utf8]{inputenc} % allow utf-8 input
\usepackage[T1]{fontenc}    % use 8-bit T1 fonts
\usepackage{hyperref}       % hyperlinks
\usepackage{url}            % simple URL typesetting
\usepackage{booktabs}       % professional-quality tables
\usepackage{amsfonts}       % blackboard math symbols
\usepackage{nicefrac}       % compact symbols for 1/2, etc.
\usepackage{microtype}      % microtypography
\usepackage{xcolor}         % colors

\usepackage[utf8]{inputenc} % allow utf-8 input
\usepackage[T1]{fontenc}    % use 8-bit T1 fonts
\usepackage{hyperref}       % hyperlinks
\usepackage{url}            % simple URL typesetting
\usepackage{booktabs}       % professional-quality tables
\usepackage{amsfonts}       % blackboard math symbols
\usepackage{nicefrac}       % compact symbols for 1/2, etc.
\usepackage{microtype}      % microtypography
\usepackage{xcolor}         % colors

\title{Conditional Density Estimation with Histogram Trees}
\author{%
  Lincen Yang \quad \quad \quad \quad Matthijs van Leeuwen \\
  LIACS, Leiden University\\
  Einsteinweg 55, 2333CC Leiden, The Netherlands
 \\
  \texttt{\{l.yang, m.van.leeuwen\}@liacs.leidenuniv.nl} \\    
}

\begin{document}

\maketitle

\begin{abstract}
  % The abstract paragraph should be indented \nicefrac{1}{2}~inch (3~picas) on
  % both the left- and right-hand margins. Use 10~point type, with a vertical
  % spacing (leading) of 11~points.  The word \textbf{Abstract} must be centered,
  % bold, and in point size 12. Two line spaces precede the abstract. The abstract
  % must be limited to one paragraph.
  % Conditional density estimation (CDE) goes beyond regression by modeling the full conditional distribution instead of just the conditional mean, making it particularly useful when the data is heteroscedastic or multi-modal, or when the goal is understanding the data besides prediction. Although the latter goal calls for interpretability, interpretable CDE methods have been understudied. 
    % Specifically, existing methods typically build upon kernel density estimation, whereas tree-based methods are arguably more comprehensible due to their conceptual simplicity and suitability for visualization. 
  % Last, we explore using CDTree for visualizing the distribution of the medical costs conditioned on the patients' demographic features. 
  % and hence provides more information for understanding the data than the conditional mean in regression. 

    % due to the conceptual simplicity and suitability for visualization, tree-based methods are arguably more comprehensible, yet disregarded for the CDE task. 

  Conditional density estimation (CDE) goes beyond regression by modeling the full conditional distribution, providing a richer understanding of the data than just the conditional mean in regression. This makes CDE particularly useful in critical application domains. However, interpretable CDE methods are understudied. Current methods typically employ kernel-based approaches, using kernel functions directly for kernel density estimation or as basis functions in linear models. In contrast, despite their conceptual simplicity and visualization suitability, tree-based methods---which are arguably more comprehensible---have been largely overlooked for CDE tasks. Thus, we propose the Conditional Density Tree (CDTree), a fully non-parametric model consisting of a decision tree in which each leaf is formed by a histogram model. Specifically, we formalize the problem of learning a CDTree using the minimum description length (MDL) principle, which eliminates the need for tuning the hyperparameter for regularization. Next, we propose an iterative algorithm that, although greedily, searches the optimal histogram for every possible node split. Our experiments demonstrate that, in comparison to existing interpretable CDE methods, CDTrees are both more accurate (as measured by the log-loss) and more robust against irrelevant features. Further, our approach leads to smaller tree sizes than existing tree-based models, which benefits interpretability. 
  
\end{abstract}

\section{Introduction} \label{sec:intro}
Conditional density estimation (CDE) is a crucial yet challenging task in modeling the associations between the features and the \emph{continuous} target variable, which has received a lot of research interest since the 1970s~\citep{rosenblatt1969conditional}. By modeling the full conditional distribution, CDE is useful when the datasets are multi-modal, heavily skewed, or heteroscedastic. As a result, it is widely applied in various fields, including Genome~\citep{desantis2014breast}, Astronomy~\citep{ball2008robust}, wind power forecasting~\citep{jeon2012using}, and computer network~\citep{samani2021conditional}. 

% As CDE provides richer information for \emph{understanding the data} than regression, which only models the conditional mean, CDE is desirable for well-informed decision-making in critical areas. Consequently, it has received significant attention in critical application domains such as healthcare~\citep{nikolova2012maximum,strobl2021dirac,gilleskie2004flexible}, which calls for interpretability.For instance, in clinical decision support, understanding \emph{why} single patients have certain conditional density estimates may be required besides having accurate estimates, as physicians may only rely on interoperable results. 
CDE provides richer information for data understanding than regression, which only models the conditional mean. This makes CDE desirable for well-informed decision-making in critical areas,  such as healthcare~\citep{nikolova2012maximum,strobl2021dirac,gilleskie2004flexible}, which calls for interpretability.

% However, recent research for CDE focuses on black-box models such as neural networks~\citep{dutordoir2018gaussian, trippe2018conditional, rezende2015variational,rothfuss2019conditional,shu2017bottleneck} and tree ensembles~\citep{gao2022lincde}, 
Despite its importance, recent CDE research has predominantly focused on black-box models, such as neural networks~\citep{dutordoir2018gaussian, trippe2018conditional, rezende2015variational,rothfuss2019conditional,shu2017bottleneck} and tree ensembles~\citep{gao2022lincde}.
In contrast, intrinsically interpretable models for CDE are understudied. Specifically, decision tree-based methods are largely neglected, with CADET~\citep{cousins2019cadet} being the only existing method to the best of our knowledge.
% Particularly, methods based on decision trees are largely neglected: to the best of our knowledge, the only CDE method based on the decision tree is CADET~\citep{cousins2019cadet} (although several black-box tree ensemble models also exist~\cite{gao2022lincde,pospisil2018rfcde}). 
However, CADET's assumption of a Gaussian distribution for the target variable in each leaf node limits its ability to model complex conditional densities.
% However, CADET assumes a Gaussian distribution for the target variable conditioned on each leaf node, which is rather restrictive and limits its ability to model the full conditional density when the ``true" conditional density function is multi-modal or skewed.

As a result, kernel-based models became the standard among `shallow' models for CDE, including methods based on kernel density estimation (KDE)~\citep{rosenblatt1969conditional}, and linear models with the basis functions chosen as Gaussian kernels~\citep{sugiyama2010conditional,fan1996estimation}. Nevertheless, kernel-based models are arguably less interpretable than decision trees: as conditions in decision trees are directly readable, they are comprehensible to humans without statistical expertise (e.g., individuals affected by data-driven decisions rather than professional data analysts). 

\begin{figure}[ht]
    \centering
    \includegraphics[width=\textwidth]{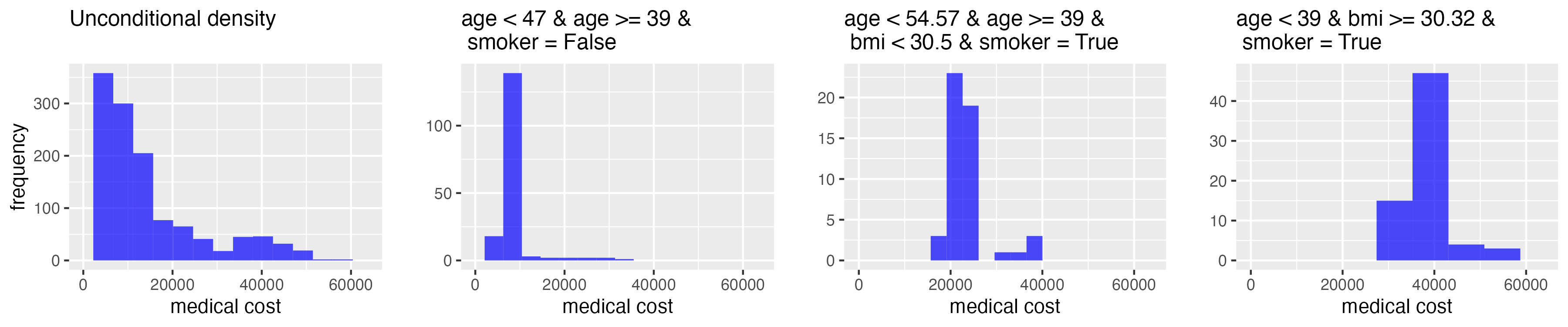}
    \caption{Three selected leaves from the CDTree modeling the conditional density of the medical costs given demographic features, together with the unconditional density for medical costs.}
    \label{fig:medical}
\end{figure}

% Thus, to introduce a flexible tree-based CDE model and hence overcome the shortcomings of CADET, we propose the Conditional Density Tree (CDTree), in which each leaf consists of a histogram model. 
To address these limitations, we propose the Conditional Density Tree (CDTree), a flexible tree-based CDE model in which each leaf node consists of a histogram model.
For illustration, Figure~\ref{fig:medical} shows the CDTree learned from a dataset about the personal medical cost with demographic features~\citep{lantz2019machine}. The figure visualizes the histogram models for the conditional densities on three selected leaf nodes, along with the unconditional density. 
The readable rules, extracted from the tree paths, and the histogram visualizations, make the results easily comprehensible to humans without statistical expertise. For instance, the difference in medical costs between smokers and non-smokers is evident when comparing the plots. The full results with histograms on all leaf nodes, as well as further descriptions of the dataset, are provided in Appendix~\ref{appendix:medical}.

Learning a CDTree from data is a challenging task, as it requires simultaneously optimizing the decision tree structure and the number of bins for histograms on all leaf nodes. While often used for either task, cross-validation is too time-consuming if we need to search for the optimal number of bins for every possible node split. 
Thus, we formalize the learning problem using the minimum description length (MDL) principle~\citep{grunwald2019minimum,rissanen1978modeling}, which we briefly review in Section~\ref{subsec:mdl_preliminary}. Adopting MDL eliminates the need for tuning the hyperparameter, by cross-validation, for both regularizing the decision trees~\citep{breiman2017classification} and for choosing the number of bins for histograms. 

Our main contributions are as follows. First, we introduce the CDTree for the CDE task and formalize the learning problem with the MDL principle. Second, we propose an iterative algorithm that can search the optimal histogram for all possible node splits. Third, we benchmark against a wide range of competitors and demonstrate that CDTree is highly competitive. Specifically, CDTree is more accurate than existing interpretable CDE methods (as measured by the log-loss). Meanwhile, CDTree has smaller tree sizes than other tree-based methods, which benefits interpretability. In addition, CDTree is extremely robust to irrelevant features, which is noteworthy as irrelevant features are known to harm the convergence rate for CDE~\citep{hall2004cross}. Further, we argue that the (intrinsic) explanations of CDTree are trustworthy only if the CDTree is robust against irrelevant features. 

\section{Related Work}\label{sec:related}
% \textbf{KDE-based CDE.} Several CDE methods have been developed based on kernel density estimation (KDE). The most straightforward approach, often referred to as conditional-KDE (CKDE)~\citep{rosenblatt1969conditional}, is to separately estimate the joint and marginal densities by (unconditional) KDE, and then use their ratio for CDE. Additionally, the $\epsilon$-neighborhood KDE (NKDE) estimates the conditional density $f(y|x)$ by applying KDE on the subset of data points within the neighborhood of the point to be estimated, with the neighborhood range controlled by the parameter $\epsilon$.
\textbf{KDE-based CDE.} Several CDE methods have been developed based on kernel density estimation (KDE). The most straightforward approach, known as conditional-KDE (CKDE)~\citep{rosenblatt1969conditional}, involves separately estimating the joint and marginal densities using unconditional KDE and then taking their ratio to obtain the conditional density. Another approach, $\epsilon$-neighborhood KDE (NKDE), estimates the conditional density by applying KDE to the subset of data points within the $\epsilon$-neighborhood of the target point, with the range of the neighborhood controlled by the parameter $\epsilon$.

% However, KDE-based methods suffer from several shortcomings. First, KDE is arguably less interpretable than tree-based models, as comprehending KDE requires statistical knowledge, making it less accessible to domain experts in other fields and the general public. Furthermore, while decision trees can naturally handle both discrete and continuous feature values, discrete feature variables can cause considerable issues for kernel-based methods. Specifically, CKDE involves estimating the joint density for the feature and target variable; hence, when features contain discrete variables, discrete kernels, which are more difficult to interpret~\citep{mussa2013aitchison}, are required. 
However, KDE-based methods have several limitations. First, KDE is arguably less interpretable than tree-based models, as understanding KDE requires statistical knowledge, making it less accessible to domain experts and the general public. Additionally, while decision trees can naturally handle both discrete and continuous feature values, discrete features pose significant challenges for kernel-based methods. Specifically, CKDE requires estimating the joint density of the features and target variable, which necessitates the use of discrete kernels for discrete variables. These discrete kernels are often difficult to interpret~\citep{mussa2013aitchison}. As for NKDE, when both discrete and continuous feature values exist, the choice of the scale for the continuous variables unavoidably introduces a certain degree of arbitrariness in defining the distances used to characterize the neighborhood.

\textbf{Regression-based CDE.} Motivated by several issues of CKDE, including high variance as a plug-in estimator, the exponentially growing search space for bandwidth tuning, and the curse of dimensionality for estimating the joint density function, the method named least-squares CDE (LSCDE)~\citep{sugiyama2010conditional} was proposed. LSCDE aims to directly estimate the ratio between the joint and marginal densities by assuming this ratio is a linear combination of several basis functions, chosen to be Gaussian kernels. Similarly, \citet{fan1996estimation} proposed double-kernel local linear regression, which transforms the conditional density function into the conditional expectation of the (unconditional) density function of the target variable, leading to a least-squares approach as well. However, both methods face the challenge of bandwidth selection, as the value of the Gaussian kernel function must be calculated with the \emph{entire} feature vector as input. This issue becomes particularly problematic as the search space for the bandwidth grows exponentially with the number of features.

% However, in both methods, the value of the Gaussian kernel function on the entire feature vector needs to be calculated, which raises the issue of bandwidth selection, of which the search space grows exponentially as the number of features. 

% \paragraph{Tree-based CDE.} The only existing CDE method based on single trees that we are aware of is CADET~\citep{cousins2019cadet}, which improves CART by designing a node-splitting heuristic specifically for CDE. However, CADET assumes a Gaussian distribution for the target variable on each leaf, which is far less flexible than our non-parametric histogram models. Further, we also mention the related tree ensemble models for CDE. First, RFCDE~\citep{pospisil2018rfcde} adopts a two-step approach: 1) fitting a standard random forest for regression to calculate the ``weight" of each training point, and 2) estimating the CDE by the weighted average of unconditional density estimation of the target using KDE. Second, the more recently proposed LinCDE~\citep{gao2022lincde} method that learns a boosted tree model based on Lindsey's method~\citep{lindsey1974comparison}, in which each tree is equipped with a parametric (though flexible) model. While tree ensembles are often considered more powerful than single trees, we argue that it is still worthwhile to study models based on single trees for interpretability reasons. 
\textbf{Tree-based CDE.} The only existing CDE method based on single trees that we are aware of is CADET~\citep{cousins2019cadet}, which improves CART regression trees~\citep{breiman2017classification} by designing a node-splitting heuristic specifically for CDE. However, CADET assumes a Gaussian distribution for the target variable on each leaf, which is far less flexible than our non-parametric histogram models. 

% \paragraph{Black-box CDE.} Neural networks have been shown to perform well in a wide range of tasks. For CDE, the most well known methods include NF~\citep{rezende2015variational} and MDN~\citep{bishop1994mixture}. Further, we mention the tree ensemble models that are conceptually closer to our CDTree. To begin with, RFCDE~\citep{pospisil2018rfcde} adopts a two-step approach by firstly fitting a standard random forest, and then estimating the CDE by the weighted average of the unconditional density estimates obtained by KDE, in which the weights are calculated based on the random forest. Furthermore, the more recently proposed LinCDE~\citep{gao2022lincde} method learns a boosted tree model based on Lindsey's method~\citep{lindsey1974comparison}, in which each tree is equipped with a parametric model. While black-box models are often accurate, we argue that interpretable CDE methods are valuable in critical application domains and for understanding the data.
\textbf{Black-box CDE.} Neural networks have been shown to perform well in a wide range of tasks. For CDE, the most well-known methods include NF~\citep{rezende2015variational} and MDN~\citep{bishop1994mixture}.  As for tree ensemble models, RFCDE~\citep{pospisil2018rfcde} first fits a standard random forest, and then estimates the CDE by the weighted average of the unconditional density estimates obtained via KDE, with weights determined by the random forest. Furthermore, the recently proposed LinCDE~\citep{gao2022lincde} method learns a boosted tree model based on Lindsey's method~\citep{lindsey1974comparison}. While black-box models are highly accurate, we argue that interpretable CDE methods are valuable for applications in critical domains and for understanding the data.

\section{Conditional Density Tree with Histograms}
Tree-based models divide the feature space into disjoint (hyper-)boxes by recursively splitting on individual feature variables (for which we consider binary splits only). Consequently, the induced partition with $K$ subsets (leaves) can be represented as $M = \{S_k\}_{k \in [K]}$, where each leaf $S_k$ represents a subset of the feature space and $[K] := \{1, ..., K\}$. Further, each leaf is equipped with a single (unconditional) density estimator, denoted as $f_k(.)$. Thus, for a dataset $D = (x^n, y^n)$ with sample size $n$, the tree-based model $M$ first identifies the leaf to which each $(x,y) \in D$ belongs, and then estimates the conditional density $f(y|x)$ as $f_k(y)$ (assuming $x \in S_k)$. 

We choose histograms as our model class for each $f_k(.)$. Unlike previous parametric models~\citep{gao2022lincde, cousins2019cadet}, which carry the risk of misspecification, histograms are non-parametric yet efficient. 
Additionally, in comparison to KDE, histograms do not require selecting a kernel or tuning the bandwidth.

% since histograms, as non-parametric models, prevent the risk of mis-specification while remaining both efficient and flexible. 
% Consider the subset of data $D_k := \{(x,y) \in D | x \in S_k\}$, 

% Consider the random vector $(X,Y)$ that independently `generates' the dataset $D=(x^n,y^n)$, a (fixed) histogram model partitions the domain of $Y$ into equal width bins, and hence approximates the density of $Y$ by piece-wise constants estimated from data. 
We next describe our notations and formally present the histogram as a probabilistic model. Formally, a (fixed) histogram model partitions the domain of the target variable $Y$ into equal-width bins, and then approximates the density of $Y$ by piece-wise constants estimated from data. Thus, we can denote a histogram model as a tuple $H = (B_l, B_u, h)$, with $B_l$ and $B_u$ respectively representing the lower and upper boundary of the histogram, and $h$ the number of bins. A histogram model $H$ can parameterize a family of distributions by $\alpha = (\alpha_1, ..., \alpha_h)$, with the probability density function in the form of $f_H(Y=y) = \frac{\alpha_j}{(B_u - B_l)/h}, \,\,\, \forall y \in [B_l + (j-1)\frac{B_u - B_l}{h}, B_l + j \frac{B_u - B_l}{h})$, in which $(B_u - B_l)/h$ is the bin width and $[B_l + (j-1)\frac{B_u - B_l}{h}, B_l + j \frac{B_u - B_l}{h})$ denotes the interval for the $j$th bin. In practice, the histogram boundaries $B_l$ and $B_u$ are often set based on the dataset at hand, by prior knowledge, or by the range of the values plus/minus a small constant. Meanwhile, the parameters $\alpha$ can be estimated by the maximum likelihood estimator (i.e., by the empirical frequencies in each bin):
    $\hat{\alpha}_j =\frac{1}{n} {\sum_{i=1}^n \mathbf{1}_{[B_l + (j-1)\frac{B_u - B_l}{h}, B_l + j \frac{B_u - B_l}{h})}(x_i)}$,
in which $\mathbf{1}(.)$ is the indicator function. 
% In practice, it is also common to apply a Laplacian smoothing, which simply adds some pseudo-count(s) to each bin. 

% \subsection{The minimum description length principle for model selection}

% \section{Learning Tree-based Models from Data}
\section{The MDL-optimal CDTree}\label{sec:mdl}
We briefly review the minimum description length (MDL) principle, and then formalize the problem of learning a CDTree that consists of histograms as an MDL-based model selection problem. 

\subsection{Preliminary: the MDL principle for model selection}\label{subsec:mdl_preliminary}
Rooted in information theory, the minimum description length (MDL) principle states that the optimal model is the one that can compress the data most~\citep{rissanen1978modeling,grunwald2007minimum}. Precisely, given the dataset $D = (x^n, y^n)$, the MDL-optimal model is defined as
\begin{equation}\label{eq:mdl_def}
    M^* = \arg \min_{M \in \mathcal{M}} -\log_2 P_M(y^n|x^n) + L(M), 
\end{equation}
in which $M$ is a CDTree and $\mathcal{M}$ the model class of all possible CDTrees for our learning task,  meanwhile $L(M)$ is the code length (in bits) needed to transmit the model in a lossless manner. According to the Kraft's inequality~\citep{cover:12:elements}, $L(M)$ can also be regarded as a prior probability distribution defined on the model class. 
Further, $P_M(.)$ is the so-called \emph{universal distribution}~\citep{grunwald2007minimum}:
% which we describe as follows. Intuitively, consider the model $M$ in our learning task, i.e., a CDE tree, that parameterize a family of probability distributions, denoted as $P_{M, \theta}$, where $\theta$ is the parameter vector that contains the histogram parameters (i.e., the $\alpha$) for all leaf nodes. Informally, a universal distribution $P_M(.)$, which does not depend on $\theta$, is a ``representative" distribution of the distribution family $P_{M, \theta}$, in the sense that asymptotically $P_M(.)$ fits the data $D$ equally well as the ``best" single distribution in $P_{M, \theta}$. In other words, formally, the \emph{regret} of a universal distribution $P_M(.)$, defined as $\max_{y^n} \{ \max_{\theta} \log_2 P_{M, \theta}(y^n|x^n) - \log_2 P_M(y^n|x^n)\}$, must be bounded by $n \epsilon > 0, \forall \epsilon > 0$ as $n \rightarrow \infty$~\citep{grunwald2007minimum}.
as $M$ parameterizes a family of probability distributions, denoted as $P_{M, \theta}$, $P_M(.)$ can be regarded as a ``representative" distribution such that the \emph{maximum regret}, defined as $\max_{y^n} \{ \max_{\theta} \log_2 P_{M, \theta}(y^n|x^n) - \log_2 P_M(y^n|x^n)\}$, can be bounded by $n \epsilon > 0, \forall \epsilon > 0$ as $n \rightarrow \infty$~\citep{grunwald2007minimum}, in which the maximum is defined over all possible values for $y^n$ in the domain of the target variable. Intuitively, the \emph{regret} is the difference between the log-likelihood of $P_M(y^n|x^n)$ (which does not depend on $\theta$) and the maximum log-likelihood $\max_{\theta} \log_2 P_{M, \theta}(y^n|x^n)$. For our learning task, $\theta$ is the parameter vector that contains the histogram parameters (i.e., the $\alpha$'s) for all leaf nodes. 

The MDL principle has been successfully applied to various data mining and machine learning tasks~\citep{galbrun2022minimum}, and specifically to partition-based models, including histograms and classification trees/rules~\citep{quinlan2014c4,yang2022truly,proencca2020interpretable,kontkanen2007mdl}. The MDL framework provides a principled way of regularizing model complexity without any regularization hyperparameter to be tuned. Moreover, as the Bayesian marginal distribution is one specific type of the universal distributions, the MDL-based model selection can also be regarded as a generalization of Bayesian model selection~\cite{grunwald2019minimum}.

\subsection{Normalized maximum likelihood for CDTrees}
The optimal universal distribution under the MDL framework is the so-called \emph{normalized maximum likelihood} (NML) distribution, defined as~\citep{grunwald2007minimum, shtar1987universal, grunwald2019minimum} 
\begin{equation}\label{eq:sc}
P_M(y^n|x^n) = \frac{\max_{\theta} P_{M, \theta}(y^n|x^n)} {\int_{y^n} \max_{\theta} P_{M, \theta} (y^n|x^n)},
\end{equation}
under the condition that the denominator is finite. It can be shown that the NML distribution is the only distribution that leads to the minimax regret $\min_{P_M}  \max_{y^n} \{ \max_{\theta} \log_2 P_{M, \theta}(y^n|x^n) - \log_2 P_M(y^n|x^n)\}$, in which the denominator in Eq.~\ref{eq:sc} is exactly equal to the regret~\citep{grunwald2007minimum}. 
% Note that the denominator is equal to the regret of for the NML distribution $P_M(.)$. 

The denominator in Eq.~\ref{eq:sc} is in general prohibitively expensive to compute~\citep{grunwald2007minimum,silander2018quotient,roos2008bayesian}, with a few exceptions including the cases when the probabistic model represents categorical distributions~\citep{kontkanen2007linear}, decision rules for classification~\citep{yang2022truly}, and one- and multi-dimensional histogram models~\citep{kontkanen2007mdl,marx2021estimating, yang:20:unsupervised}. We extend these previous results and prove that, for CDTrees with histogram models, the denominator (regret) is finite and is equal to the products of the regret terms of the NML distributions for one-dimensional histogram models, as shown in Proposition~\ref{prop1}. This result is useful for efficiently calculating the denominator (regret) in Eq.~\ref{eq:sc} for CDTree models, as the regret terms for histogram models are known to be equal to that of the categorical distributions~\citep{kontkanen2005analyzing,kontkanen2007linear}, for which an efficient algorithm exists with sub-linear time complexity~\citep{mononen:08:sub-lin-stoch-comp}.

\begin{proposition}\label{prop1}
Let $\theta = (\alpha^1, ..., \alpha^K)$ be the histogram parameters for histograms on all leaves, and let $\hat\theta = \arg\max_\theta P_{M, \theta} (y^n| x^n)$. Then $\int_{y^n} \max_{\theta} P_{M, \theta} (y^n|x^n) = \prod_{k \in [K]} \mathcal{R}(N_k, h_k)$, in which $\mathcal{R}(N_k, h_k)$ is the regret (denominator) of the NML distribution of the histogram model on the $k$th leaf node that contains $N_k$ data points and $h_k$ bins. 
\end{proposition}

We defer the proof to Appendix~\ref{appendix:proof} due to space limitations. 

\subsection{Code length for the model}\label{subsec:cl_model}
To encode a CDTree model in a lossless manner, we need to sequentially encode 1) the number of nodes in the decision tree, 2) the structure of the tree, 3) the splitting condition for each internal node in a predetermined order (e.g., depth first), 4) the number of bins for the histogram on each leaf node, and 5) the boundaries $B_l$ and $B_u$ of the histograms. That is, let $L(.)$ denote the function that calculates code length in bits, the code length needed to encode a model $M$ can be decomposed to 
\begin{equation}\label{eq:cl_model}
\begin{aligned}
    L(M) = & L(\text{\# leaves}) + L(\text{tree structure})  + \sum_{j \in [K - 1]} L(\text{splitting condition of the $j$th \emph{internal} node}) \\
    & + \sum_{k \in [K]} L(\text{\# bins of histogram on the $k$th \emph{leaf} node}) + L(\{B_l, B_u\});
\end{aligned}
\end{equation} 
we next describe them in order. 

\textbf{Encoding the tree size and structure.}
As we consider full binary trees only, it is sufficient to encode the number of leaves $K$, which also determines the number of total nodes. As $K$ is a positive integer, we adopt the standard Rissanen’s integer universal code~\citep{rissanen1983universal}, denoted as $L_{\mathbf{N}}(K)$, for which the code length is equal to $L_{\mathbf{N}}(K) = c + \log_2(K) + \log_2\log_2(K) + ...$; the summation continues until a small enough precision is reached, and $c \approx 2.865$ is a constant.

Further, for full binary trees, the number of all possible tree structures for trees with $K$ leaves is equal to the Catalan number, denoted as $\mathcal{C}_{K-1}$~\citep{stanley2015catalan}. Hence, specifying one certain structure costs $\log_2 \mathcal{C}_{K-1}$ bits. Thus, $L_{\mathbf{N}}(K) + \log_2 \mathcal{C}_{K-1}$ bits are required for encoding the tree size and structure.

% \consider{Notably, while this is not the only option for encoding the tree size and structure \textbf{\textcolor{red}{[??]}}, the encoding scheme we adopted rigorously follows the guideline of achieving the so-called conditional minimax defined in \textbf{\textcolor{red}{[??]}}. }
Note that while there exist multiple ways of encoding tree size and structure (e.g., one alternative is to leverage the joint probability of the tree size and the tree structure, which can be defined by treating the tree as a realization of a Galton-Watson process~\citep{papageorgiou2024posterior, athreya2004branching}), our encoding scheme  adheres to the principle of achieving \emph{conditional minimax}~\citep{grunwald2019minimum} by explicitly putting a prior on the number of nodes. 

\textbf{Encoding the splitting conditions.}
% We propose an encoding scheme that does not require the hyperparameter for specifying the resolution or data precision for encoding the splitting conditions for continuous variables, as the choice for the resolution may be a bit arbitrary in practice~\citep{kontkanen2007mdl,proencca2020interpretable,yang:20:unsupervised,yang2022truly,marx2021estimating}. Instead, we integrate the required resolution for continuous variables as part of the model itself. 
An individual splitting condition on a single tree node consists of a variable name and a splitting value, which we encode sequentially.
% and we encode the name and the value sequentially. 

% We encode an individual splitting condition by firstly specifying which variable the condition contains, and then specifying the corresponding value. 
To begin with, if the dataset contains $m$ feature variables, encoding the name of a certain variable $X$ costs $\log_2{m}$ bits. Further, the code length needed for encoding the splitting value depends on the variable type. First, for a discrete variable $X$ with $|\mathcal{X}|$ unique values, specifying a single value cost $\log_2(|\mathcal{X}|)$ bits. Second, encoding the splitting value for a continuous variable can be achieved by sequentially encoding 1) the granularity level for the search space, denoted as the positive integer $d$, and 2) the exact value within the granularity level $d$. Specifically, with a fixed $d$, we consider as the search space the $C\cdot 2^{d-1}$ quantiles that can partition the values of $X$ into equal-frequency bins. Note that the values of $X$ are based on the subset of data points contained in this internal node locally. 

Hence, encoding $d$ costs $L_{\mathbf{N}}(d)$ bits with the Rissanen’s code~\citep{rissanen1983universal}, and encoding one specific splitting value (quantile) costs $\log_2(C\cdot 2^{d-1}) =\log_2(C) + d - 1$ bits. That is, we treat the granularity level $d$ as a parameter to be optimized when learning a CDTree from the data, which avoids (arbitrarily) specifying the granularity in advance, a shortcoming in previous MDL-based methods (for other tasks instead of CDE)~\citep{kontkanen2007mdl,proencca2020interpretable,yang:20:unsupervised,yang2022truly,marx2021estimating}. In contrast, the parameter $C$ can be used to express the prior belief about the hierarchical structure of the search space, as further discussed in Appendix~\ref{sub_append:cdtree_parameters}.

\textbf{Encoding the histograms.} One subtle choice we made is to set the boundaries for histograms on all leaves to be same, and we set them as the global boundary of the target variable. This avoids unseen (test) data points falling outside the boundaries of the histograms on the leaf nodes. This is because while it may be common to assume that the boundaries for the histogram are known in \emph{unconditional} density estimation, assuming the same for CDE is hardly realistic. 

Therefore, the code length needed to encode the boundary $L(\{B_l, B_u\})$ in Eq.~\ref{eq:cl_model} is a constant that does not affect the model selection result. Consequently, it suffices to encode the number of bins for each histogram: for the histogram on the $k$th leaf with $h_k$ bins, it costs $L_{\mathbf{N}}(h_k)$ bits by the Rissanen’s integer code~\citep{rissanen1983universal}.

% Specifically, this requires assuming that the boundary of the values for $Y$ \emph{conditioned on any event in the joint probability space} is known, which is hardly realistic in practice. 

% That is, despite the fact that assuming the boundary for the marginal distribution of $Y$ is known is common in practice~\citep{kontkanen2007mdl}, we cannot assume the boundary of variable $Y$ conditioned on any \emph{event} in the joint probability space defined on the feature and the target variable is known. 
% The rationale is to make the histogram model robust and the estimates of boundaries \emph{not} sensitive to small changes on the minimum and maximum values of data points on a single leaf. Further, this also avoids unseen (test) points to fall outside the boundary of the histogram: while it may be reasonable (and common for histogram models) to assume the boundary of the target variable $Y$ is known, it is hardly justifiable to assume that the boundary (or support) of variable $Y$ conditioned on any \emph{event} in the joint probability space of the feature and the target variable is always known. 
% By the Rissanen’s integer universal code~\citep{rissanen1983universal}, it costs $L_{\mathbf{N}}(h_k)$ bits to encode the number of bins for the histogram on the $k$th leaf. 

\subsection{Model selection criterion}
Combining Eq.~\ref{eq:mdl_def},\ref{eq:sc}, and \ref{eq:cl_model}, together with the results from Proposition~\ref{prop1}, we present the following MDL-score as our final model selection criterion:
\begin{equation} \label{eq:final_score}
    M^* = \arg\min_{M \in \mathcal{M}} - \log_2 \left(\max_\theta P_{M, \theta}(y^n|x^n)\right) + \sum_{k \in [K]} \log_2 \mathcal{R}(N_k, h_k) + L(M) 
\end{equation}
which has the form of the regularized maximum likelihood, yet without the need to tune the regularization hyperparameter. Note that the exact form of $L(M)$ depends on the types of the feature variables; e.g., by assuming all variables are continuous, we obtain 
% \begin{equation}
% \begin{split}
%     M^* & = \arg\min_{M \in \mathcal{M}} - \log \left(\max_\theta P_{M, \theta}(y^n|x^n)\right) + \sum_{k \in [K]} \log \mathcal{R}(N_k, h_k) + L_\mathbf{N}(K) + \log \mathcal{C}_{K-1} + \\
%     & + \sum_{j \in [K]} \left( \log(m) + L_\mathbf{N}(d(j)) + \log(C) + d(j) - 1 \right) + \sum_{k \in [K]}L_\mathbf{N}(h_k),
% \end{split}
% \end{equation}
\begin{equation*}
    L(M) = L_\mathbf{N}(K) + \log_2 \mathcal{C}_{K-1} + \sum_{j \in [K-1]} \left( \log_2(m) + L_\mathbf{N}(d_j) + \log_2(C) + d_j - 1 \right) + \sum_{k \in [K]}L_\mathbf{N}(h_k),
\end{equation*}
where, as defined previously, $\mathcal{R}(.)$ denotes the regret term of each histogram, $L_\mathbf{N}(.)$ the Rissanen’s integer universal code~\citep{rissanen1983universal}, $m$ the number of feature variables, $d_j$ the granularity level for the search space for the splitting values corresponding to the $j$th internal node, $C$ the constant that controls the hierarchical structure of the search space of the splitting values, and $h_k$ the number of bins for the histogram on the $k$th leaf node.

\section{Algorithm} \label{sec:algorithm}
While finding the optimal tree-based model with the branch-and-bound approach is possible for regression~\citep{zhang2023optimal} and classification~\citep{hu2019optimal}, it does not apply to our task for two reasons. First, our MDL-based regularization term differs from traditional penalty terms based on tree size. Second, our task resembles optimizing \emph{model trees}~\citep{malerba2004top,quinlan1992learning} rather than classification and regression trees, as we aim to find the optimal histogram for each candidate split. 
\begin{algorithm}[ht]
\caption{Learn CDTree from data}\label{alg:cde}
\KwIn{Training dataset $D$}
\KwOut{CDTree $M$}
\BlankLine
$M \leftarrow \{S_0\}$ \tcp*{One leaf node only}
\While{True}{
\For{$S \in M$}{ 
Search the condition that splits $S$ into two nodes and minimizes the MDL-score (Eq.~\ref{eq:final_score}). \tcp*{Described in detail in Algorithm~\ref{alg:find_split}}
}
\eIf{Splitting any $S \in M$ cannot further decrease the MDL-score}{
\Return{CDTree $M$}
}{
Among all $S \in M$, find the single $S^*$ to be split that minimizes the MDL-score.

Update $M$ by replacing $S^*$ with its two child nodes that minimize the MDL-score.
}
}
\end{algorithm}

\noindent \textbf{A greedy approach for tree construction.} We thus take a heuristic approach to optimize our MDL-score in Eq.~\ref{eq:final_score}. 
As summarized in Algorithm~\ref{alg:cde}, we start with a tree with one leaf node $M=\{S_0\}$; next, we iteratively update $M$ by replacing one of the leaf nodes with its `best' two child nodes. 
Specifically, to achieve the lowest MDL-score at each iteration, we simultaneously search for 1) which node to split, 2) the splitting condition for that node, and 3) the optimal number of bins for the histograms. 
That is, we iterate over all leaf nodes in $M$; for each leaf node, we search for the splitting condition, along with the number of bins for the histograms on the (potential) child nodes, which as a whole minimizes the MDL-score. Notably, while an exhaustive search for the `best' models on all potential child nodes is considered infeasible in traditional \emph{model tree} methods~\citep{malerba2004top,quinlan1992learning}, we empirically demonstrate in Section~\ref{subsec:exp_runtime} that our algorithm is comparable to KDE-based methods. 

\noindent \textbf{Finding the child nodes.} We next elaborate on the search for the tree-splitting condition at each iteration, for which the pseudo-code is provided in Algorithm~\ref{alg:find_split}. For simplicity, we assume all feature variables are either continuous or binary (i.e., categorical features are one-hot encoded in our implementation).

Further, we iterate over all columns of the feature matrix. For each column, we start with the granularity level $d=1$ and generate candidate split points as the $C \cdot 2^{d-1}$ quantiles that lead to equal frequency binning of the values. Note that the quantiles are generated based on the subset of data points covered by the node to be split, rather than the entire dataset. We search for the best split point at the fixed $d$ that yields the minimum MDL score among all generated candidates. We then proceed to the next granularity level $d+1$ and repeat the process, which stops if no split point in the next granularity level results in a better (smaller) MDL score. 
% Specifically, in each iteration, we go over all the existing leaf nodes in $M$, split each leaf node to find the child nodes that minimize the MDL-score, and pick the node to split again based on the MDL-score should we replace it by its children. 

% For a single node, 

% Specifically, to achieve the lowest MDL-score at each iteration, we simultaneously search for 1) which node to split, 2) the splitting condition for that node, and 3) the optimal number of bins for the histograms. 
% That is, we iterate over all leaf nodes in $M$; for each leaf node, we search for the splitting condition, along with the number of bins for the histograms on the (potential) child nodes, which as a whole minimizes the MDL-score. 

% Last, our algorithm stops when the minimum MDL-score is reached, eliminating the need for tree pruning. As we avoid creating a large tree (to be pruned) in the first place, the sizes of trees learned by our algorithm are smaller than other tree-based methods and hence benefit interpretability, which will be demonstrated in Section~\ref{subsec:tree_size}.

% We provide further specifics and the pseudo-code, in Appendix~\ref{appendix:algs}, in which we elaborate on the search for 1) the tree-splitting condition at each iteration, and 2) the number of histogram bins. 
Last, we also need to search for the optimal number of bins for the histogram that leads to the minimum MDL score, which we defer to Appendix~\ref{appendix:algs}.

\begin{algorithm}[ht]
\caption{Find the best split for node $S$} \label{alg:find_split}
\KwIn{Training set $D_S = (x^{\{S\}}, y^{\{S\}})$ covered by node $S$, with $x^{\{S\}} = (\vec{x}_1, ..., \vec{x}_m)$ having $m$ columns, and constant $C$} 
% \tcp{Without loss of the generality, assume both continuous/binary are continuous to simplify the notations. }
\KwOut{The splitting condition of $S$ that minimizes the MDL-score}
\BlankLine
\For{$j \in \{1,2,...,m\}$}{
$d \leftarrow 1$

\While{True}{
$candidate\_splits \leftarrow $ $\text{The} \,\, C \cdot 2^{d-1}$ $\text{quantiles}$ for equal-frequency binning for $\vec{x}_j$

\For{$s \in candidate\_splits$}{
    $D_l \leftarrow \{(x,y) \in D_S | x_j \leq s\}$ \tcp*{Always $s = 1/2$ for binary features}
    
    $D_r \leftarrow \{(x,y) \in D_S | x_j > s\}$ 

    Construct histograms for $D_l$ and $D_r$ that minimize the MDL-score conditioned on the fixed $j$ and $d$. \tcp*{Described in detail in Algorithm~\ref{alg:hist}} 

    Score $\leftarrow$ Calculate the MDL-score assuming $D_S$ is split into $D_l$ and $D_r$
}

    \lIf{The best Score is worse than that of the previous d}{
    \textbf{Break; else} $d \leftarrow d + 1$
    }
    % \lElse{
    % $d \leftarrow d + 1$
    % }
}
Record the best tuple $(d,s)$ for this column index $j$
}
\Return{The tuple $(j, d, s)$ that minimize the MDL-score}
\end{algorithm}

\section{Experiments}\label{sec:exp}
% We next present our experiments to demonstrate the empirical performance of CDTree in various perspectives. Specifically, we aim to answer the following research questions: 1) Does the CDTree learned from data has more accurate conditional density estimation in comparison to existing interpretable (tree-based and kernel-based) methods? 2) As a proxy for interpretability, how are the sizes of the CDTrees in comparison to other tree-based CDE models? 3) Is CDTree robust against ``noisy" features that are known to be irrelevant to the target variables? 4) Is the runtime of our algorithm for learning CDTree comparable to kernel-based methods?
% We present our experiments to demonstrate the empirical performance of CDTree from various perspectives. Specifically, we aim to answer the following research questions: 1) Does CDTree provide more accurate conditional density estimation compared to existing interpretable (tree-based and kernel-based) methods? 2) How are the sizes of CDTrees as a proxy for interpretability, in comparison to other tree-based models? 3) Is CDTree robust against irrelevant "noisy" features? 4) Are the runtimes of our algorithm comparable to those of kernel-based methods?
We present our experiments to demonstrate the empirical performance of CDTree from various perspectives. Specifically, we aim to answer the following research questions: 1) Does CDTree provide more accurate conditional density estimation compared to existing interpretable (tree-based and kernel-based) methods? 2) As a proxy for interpretability, does CDTree has smaller tree size in comparison to other tree-based models? 3) Is CDTree robust against irrelevant "noisy" features? 4) Are the runtimes of our algorithm comparable to those of kernel-based methods?

\subsection{Experiment setup}
% We use 14 datasets with numerical target variables from the UCI repository~\footnote{The UCI Machine Learning Repository\url{https://archive.ics.uci.edu}
% }. These datasets varies significantly on their source domains, sample sizes, and number of features; the latter two are summarized in Table~\ref{table:data_info}. 
% To benchmark performance, we compare our method, CDTree, to various competitor methods for conditional density estimation, using five-fold cross-validation for all reported results. 

We use 14 datasets with numerical target variables from the UCI repository~\citep{uci_link}. These datasets, summarized in Table~\ref{table:data_info}, cover a wide range of sample sizes and dimensionalities. We benchmark our method against a variety of competitors, with all results obtained on the test sets using five-fold cross-validation. Our competitors include 1) NKDE and CKDE~\citep{rosenblatt1969conditional,rothfuss2019conditional}, which are based on kernel density estimation, with bandwidth and $\epsilon$ (for NKDE only) tuned by cross-validation; 2) LSCDE~\citep{sugiyama2010conditional}, which directly estimates the ratio of joint and marginal density using linear models with Gaussian kernel basis functions; 3) the tree-based method CADET~\citep{cousins2019cadet}, which fits a Gaussian distribution on each leaf node; 4) two more tree-based baselines introduced by us, CART-k and CART-h, which respectively fits a KDE model and a histogram after a CART regression tree~\citep{breiman2017classification} is learned from data. 
Furthermore, we compare against three black-box models as "upper" baselines, including 1) neural network methods NF~\citep{rezende2015variational} and MDN~\citep{bishop1994mixture}, which apply both dropout and noise regularization~\citep{rothfuss2019noisereg}, and 2) the recently proposed tree boosting method LinCDE~\citep{gao2022lincde}. Nonetheless, our goal with CDTree is \emph{not} to be more accurate than black-box models, but to introduce a CDE method that is both interpretable and accurate.
\input{data_info_tab}

For reproducibility, we provide further details about implementations and parameter choices in Appendix~\ref{appendix:exp_setup}. We made our source code public: \url{https://github.com/ylincen/CDTree}. 

% As our main goal with CDTree is to introduce an accurate \emph{interpretable} model for CDE
% Nonetheless, our main goal with CDTree is \emph{not} to introduce the most accurate CDE models, but to provide a tree-based model for CDE that is both accurate and interpretable.

% Note that the comparison between CDTree and CART-h also servers as an ablation study for the necessity of including the MDL-histograms in the training phase in our proposed method. 

\subsection{Conditional density estimation accuracy}
In Table~\ref{table:cde_loglikelihood}, we report the average negative log-likelihoods (NLL) on the test sets, 
which can be regarded as an approximation to the expected log-loss $E(-\log(y|x))$, 
the standard loss function for measuring CDE accuracy.

\input{exp1_all_tab}

\smallskip \noindent \textbf{Comparison to interpretable models.}
The NLL of the CDTree models are the best (lowest) in 6 out of 14 datasets among all interpretable models, which are shown in bold in the table. Further, among all interpretable models, our average rank is the best, as reported in the (last row) of Table~\ref{table:cde_loglikelihood}. As the datasets are \emph{increasingly} ranked based on their dimensionalities, we observe that CKDE---although also achieving the best NLL in 6 datasets---only performs well on datasets with low dimensions. Additionally, the other two kernel-based methods LSCDE and NKDE have in general worse accuracy. 
% as shown by their average ranks. 

Further, CDTree has better accuracy (lower NLL) than all tree-based competitors on almost all datasets (13 out of 14 datasets for CADET, and 12 out of 14 datasets for CART-h and CART-k). The superiority of CDTree highlights the advantages of 1) using non-parametric histogram models, and 2) conducting an exhaustive search for optimal histograms for all node splits when learning the tree structure, rather than fitting the model after the tree structure is fixed, as in CART-h and CART-k. 
% The superiority of CDTree shows the advantage of 1) the non-parametric histogram models (over the parametric model in CADET), and 2) our exhaustive search for the optimal histograms for all node splits when learning the tree structure (instead of fitting the model after the tree structure is fixed, as in CART-h and CART-k). 

% Next, the other tree-based competitor CADET has worse (larger) NLLs than our method in almost all datasets, together with anomalously large NNLs on several datasets (`forestfir', `skillcraf', and `support2').
% Due to the instability of CADET, two more tree-based baselines by respectively fitting histograms (CART-h) and kernel density estimations (CART-k) are introduced.
% As KDE requires intensive parameter tuning for bandwidth selection, it is infeasible to fit the KDE to the data for all possible tree splits. Thus, our empirical results demonstrate that fitting KDE after the tree structure is fixed will indeed lead to sub-optimal conditional density estimates. 
% Further, since the number of bins for histograms in CART-h is learned in the same way as in CDTree under the MDL framework, the superior results of CDTree over CART-h on almost all datasets demonstrates the advantages of adopting our model selection criterion specifically for CDE in learning the tree structures. 

\textbf{Comparison to black-box models.} Neural network models generally exhibit better accuracy than interpretable models, as indicated by their average ranks. However, their non-transparency limits their applicability in critical areas. We argue that studying interpretable models and introducing CDTree paves the way for developing local surrogate models~\citep{ribeiro2016should,luo2006unsupervised}, an essential approach for generating post-hoc explainability for black-box models, as no such method currently exists for CDE.

% As shown by the average rank fo all models, neural network models have in general better accuracy than all interpretable models; nevertheless, the non-transparency limits their use in critical areas. Further, we argue that studying interpretable models and introducing the CDTree paves the way for developing local surrogate models~\citep{ribeiro2016should,luo2006unsupervised} for post-hoc explainability methods for neural networks, as there exists no such method for CDE.

% However, we argue that the introduction of CDTree is worthwhile, because 1) CDE models are widely applied in critical areas (as discussion in Section~\ref{sec:intro}), and 2) interpretable models can be used as the base for developing post-hoc explainability methods for neural networks~\citep{ribeiro2016should,luo2006unsupervised}. 
% However, we argue that it is worthwhile to study intrinsically interpretable models because 1) CDE models are widely applied in critical areas (as discussion in Section~\ref{sec:intro}), and 2) interpretable models can be used as the base for developing post-hoc explainability methods for neural networks~\citep{ribeiro2016should,luo2006unsupervised}. 
Further, the performance of CDTree is surprisingly on par with that of LinCDE. Since CDTree is based on a single tree while LinCDE is a tree ensemble model, we conjecture that the on-par performance is caused by the fact that CDTree adopts non-parametric histograms whereas LinCDE takes a parametric approach (with a much more flexible model class than Gaussian though). 

We also report the standard deviations of NLL for all methods in Appendix~\ref{appendix:sd_nll}. Our method shows no anomalously large standard deviations, indicating consistent performance across different datasets.

\begin{figure}[ht]
    \centering
\includegraphics[width=0.48\textwidth]{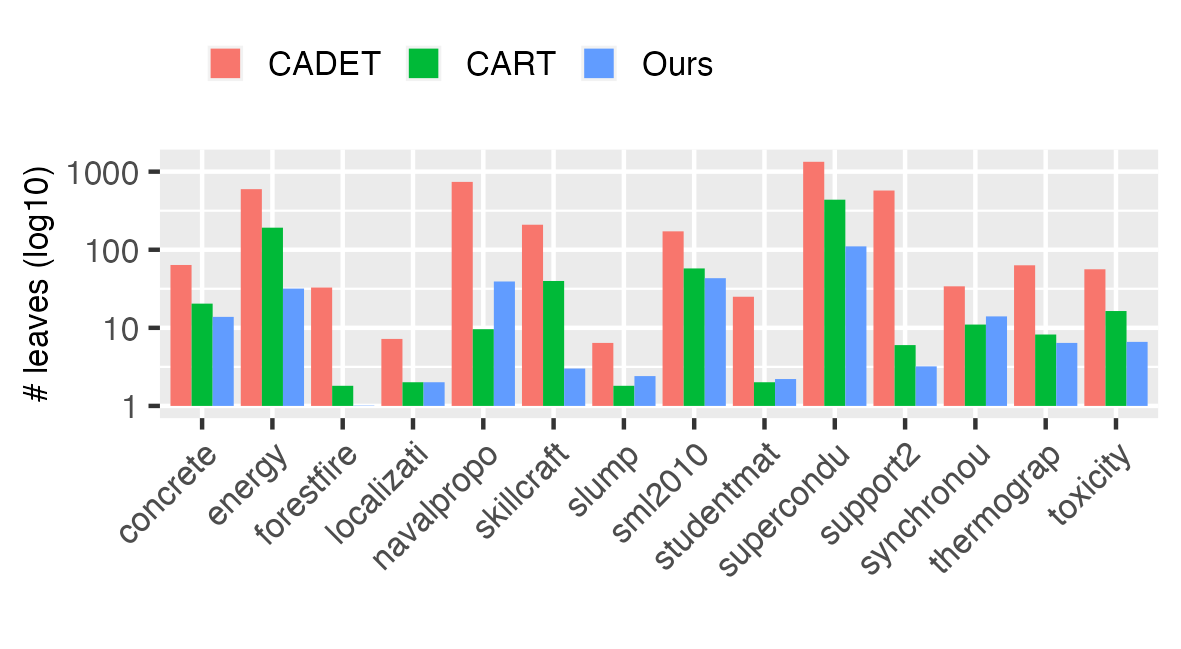}
\includegraphics[width=0.48\textwidth]{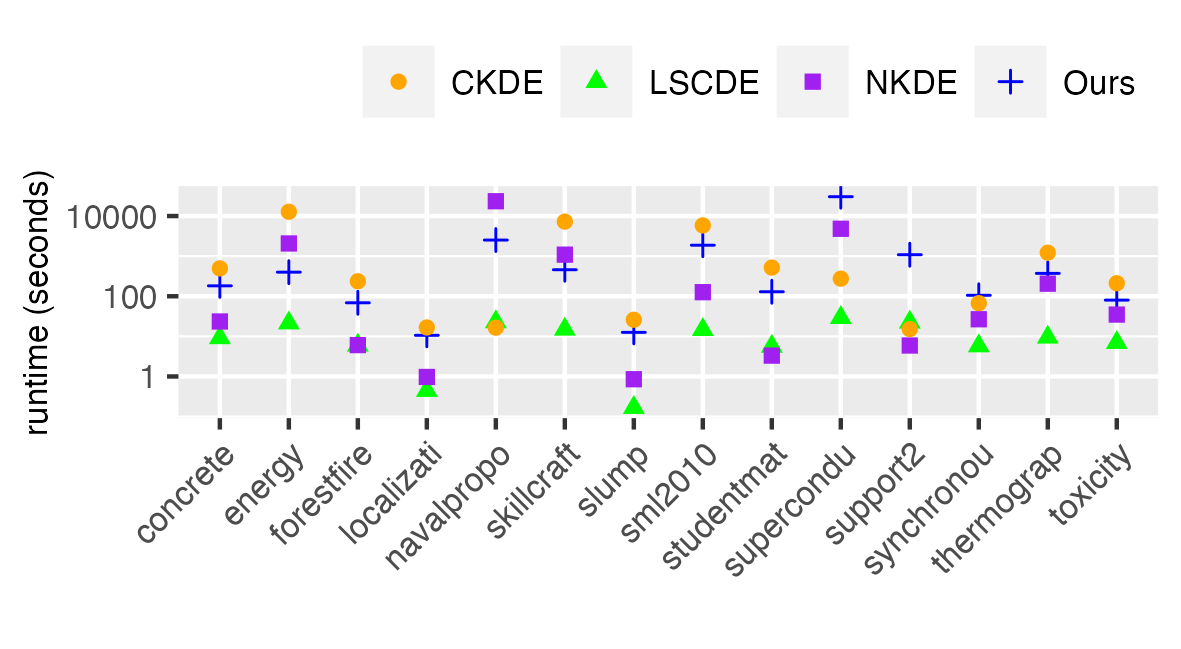}
    \caption{Left: the number of leaves for tree-based methods. Right: Runtimes of CDTree and kernel-based methods. Note that the y-axes are scaled by $\log_{10}(.)$}
    \label{fig:tree_size}
\end{figure}
\subsection{Complexity of trees}\label{subsec:tree_size}
For tree-based models, the size of the trees is an important proxy for the degree of interpretability~\citep{molnar2020interpretable}. We hence compare the number of leaves of CDTree against the other tree-based models CADET and CART (note that CART-h and CART-k have the same tree structures). We demonstrate in Figure~\ref{fig:tree_size} (left) that CDTree has smaller tree sizes than the competitors on most datasets. 

\begin{figure}[ht]
    \centering
    \includegraphics[width=\textwidth]{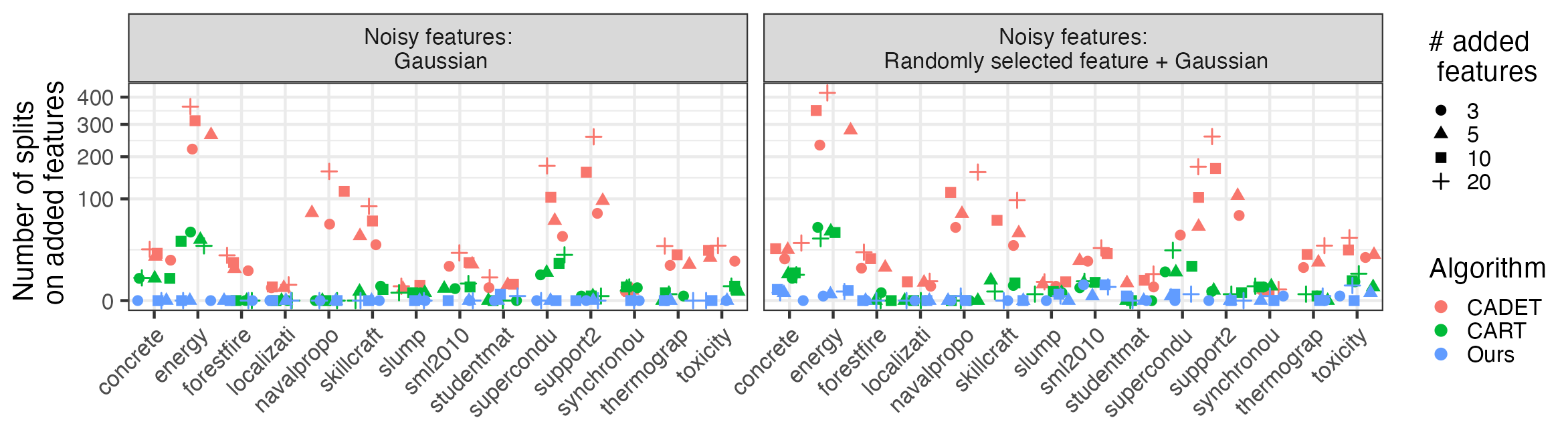}
    \caption{Number of internal nodes with split conditions that contain irrelevant features. The y-axis is scaled by the squared-root for better visualization. }
    \label{fig:wrong_splits}
\end{figure}

\subsection{Robustness to irrelevant features}
% We consider the goodness-of-fit as the number of features increases, to examine the robustness against the curse of dimensionality. We compare against NKDE and CART-k on the log-likelihood  on test datasets as the an increasing number of ``noisy" features are added to the original datasets. Specifically, the way we generate ``noisy" features are as follows.

We next investigate whether CDTree is robust against irrelevant features, which is particularly important for the interpretability of tree-based models, since the (intrinsic) explanation contained in the CDTree would not be trustworthy if such robustness did not hold. 
% A CDTree, as an intrinsically interpretable model, contains the explanation of its output in itself; however, such explanation is only trustworthy if it has high stability, in the sense that when the underlying ``true" conditional density does not change, the model (and the explanation) learned from data remains stable. 

Specifically, we generate `noisy' features in two different ways.
% two modifications to the UCI datasets which do not change the true conditional densities. 
The first way is to add $w$ irrelevant features randomly drawn from the standard Gaussian distribution to each dataset. The second way is to first randomly select $w$ features from each dataset; then, for each selected feature $X_j$, we generate a `noisy' feature by adding a Gaussian noise to it, i.e., $X_j' = X_j + N(0, s(X_j)/2)$, where $s(X_j)$ denotes the estimated standard deviation. We refer to the irrelevant features generated by the first (second) approach as independent (dependent) noisy features, where $w \in \{3,5,10,20\}$ in both cases. Note that adding $X_j'$ to the dataset does not change the true conditional density as the target variable is conditionally independent of $X_j'$ given the original feature $X_j$. 
\begin{figure}[ht]
    \centering
    \includegraphics[width=\textwidth]{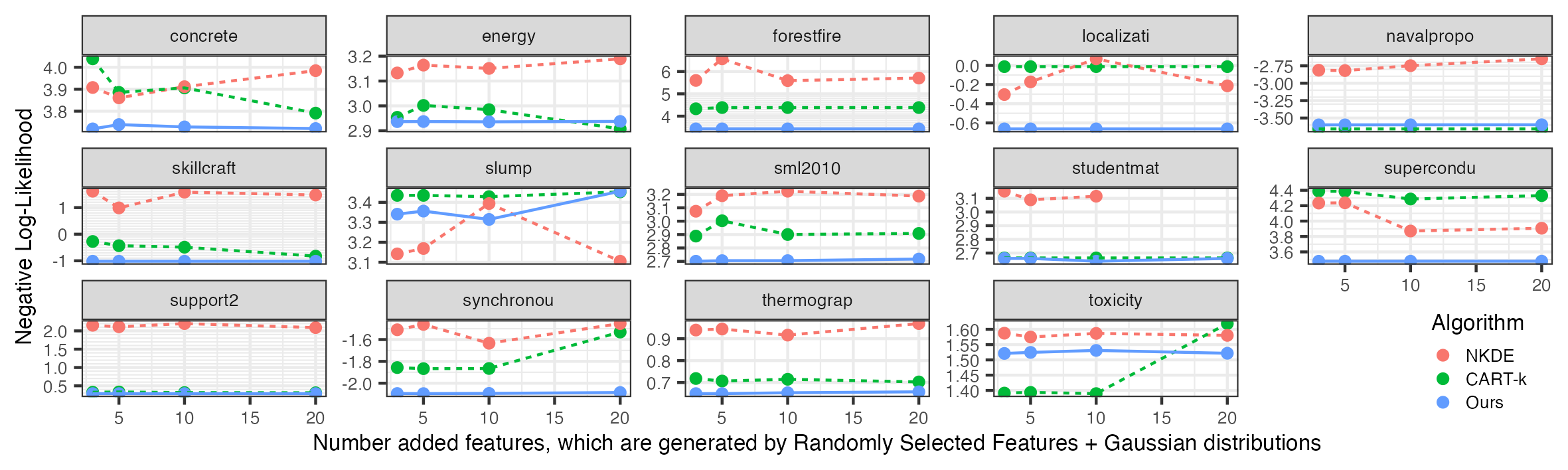}
    \caption{Negative log-likelihoods with different number of irrelevant `dependent noisy' features. The results of CDTree (shown in blue lines) are stable on all datasets.}
    \label{fig:scala}
\end{figure}
Next, we train the tree-based models on expanded datasets that include irrelevant features. We count the number of nodes with splitting conditions that involve these added irrelevant features. As demonstrated in Figure~\ref{fig:wrong_splits}, the number of splits on irrelevant features for our CDTree is almost always zero for both ways of generating noisy features. In contrast, the tree-based models learned by CART and CADET contain many more nodes with irrelevant features.

We also examine whether the negative log-likelihoods (NLL) remain stable when irrelevant features are added. For `dependent noisy' features, the results are shown in Figure~\ref{fig:scala}, where the NLL of CDTree remains nearly identical regardless of the number of added features—the only exception being the dataset "slump" (with only 103 rows). In contrast, the NLLs obtained by competitor methods are much less stable: the NLL of NKDE varies for most datasets, and CART-k shows visible changes in 6 out of 14 datasets. Similar results are observed when 'independent noisy' features are added, as shown in Appendix~\ref{appendix:scala}.
% Further, we examine whether the negative log-likelihoods (NLL) remain unchanged when irrelevant features are added. For `dependent noisy' features, we show the results in Figure~\ref{fig:scala}, in which the NLL of CDTree are almost identical for different numbers of added features---the only exception is for the dataset ``slump" (with 103 rows only). In contrast, the NLL obtained by competitor methods are much less stable: the NLL of NKDE varies for most datasets, meanwhile the results of CART-k have visible changes in 6 out of 14 datasets. Last, we also observe similar results when the `independent noisy' features are added, for which the results are shown in Appendix~\ref{appendix:scala}.

\subsection{Runtimes} \label{subsec:exp_runtime}
The idea of fitting separate models on the leaves of a decision tree exists since a long time ago~\citep{quinlan1992learning}. However, it is (still) often believed that fitting separate models for all possible node splits when growing a tree is infeasible. However, we show in Figure~\ref{fig:tree_size} (right) the runtime of CDTree and demonstrate that its runtimes are in general lower than the runtimes of CKDE (which requires intensive parameter tuning). While NKDE and LSCDE are in general faster than CDTree, the accuracy of their conditional density estimates are sub-optimal, as discussed previously. 
We exclude the comparison with other tree-based methods whose implementations are based on CART, as CART is highly optimized and known to be extremely fast. Further, as these tree-based methods either model the conditional means only (CART-h and CART-k) or assume a Gaussian model (CADET), they have far smaller search space, and hence are fast but less accurate.

% \begin{figure}[ht]
%     \centering
%     \includegraphics[width=.8\textwidth]{runtime.png}
%     \caption{Runtimes of CDTree and kernel-based methods. Note that the y-axis is scaled by $\log_{10}(.)$.}
%     \label{fig:runtime}
% \end{figure}

\section{Discussion} \label{sec:disucssion}
In this paper, we studied the interpretable conditional density estimation (CDE) models. Motivated by the fact that tree-based methods are arguably more interpretable than kernel-based methods yet have been largely disregarded for interpretable CDE, we introduced the Conditional Density Tree (CDTree). We formalized the learning problem under the MDL framework, proposed an iterative algorithm, and demonstrated its competitive empirical performance on a wide range of datasets. 

\textbf{Limitations.} As histograms are used in the CDTree, we implicitly assume that the support of the target variable is bounded. Hence, the boundaries for the histograms need to be chosen in an ad-hoc way, possibly based on prior knowledge. Further, in practice, it may happen that the unseen data points fall outside the histogram boundary, for which the predicted (conditional) density will be 0, unless the full model is re-trained. We realize that this may cause some issues when using CDTree in practice, but meanwhile, we argue that 1) this is a limitation of the histogram model itself, 2) similar issues could happen in modelling quantiles by the empirical cumulative probability function, as in well-known methods like quantile regression trees and forest, and 3)
% well-known methods like quantile regression trees and forest (as the empirical cumulative probability function cannot handle unseen data points that are larger (smaller) than the maximum (minimum) value in the training set, and 
all probability models have their own (implicit) assumptions on the probability tails, including those specifically designed for modeling the extreme value distributions~\citep{durrieu2018extremefit,kotz2000extreme}.

% \section{Case study}
% \paragraph{Spatio-temporal datasets}
% We consider the New York City Taxi Trip Duration Dataset, which was used in a Kaggle competition in 2017\footnote{\url{https://www.kaggle.com/c/nyc-taxi-trip-duration}}. 

% % An interpretable CDE model for such spatio-temporal dataset provides a lot of insights in understanding the distributions of people's departure points and destinations of their trajectories, which can be useful in distributing the Taxi cars more efficiently. 
% \begin{figure}[ht]
%     \centering
%     \includegraphics[width=\textwidth]{nyc_trip_duration_full.png}
%     \caption{Caption}
%     \label{fig:enter-label}
% \end{figure}

\section*{Acknowledgement} 
We thank the anonymous reviewers who have provided constructive and valuable reviews. We also thank Prof.dr.~Alexander Marx (TU Dortmund) for his feedback in developing the initial idea. 

\bibliography{references.bib}

\appendix
\newpage

\section{Proof}\label{appendix:proof}
\textbf{Proposition 1.}
\textit{Let $\theta = (\alpha^1, ..., \alpha^K)$ be the histogram parameters for histograms on all leaves, and let $\hat\theta = \arg\max_\theta P_{M, \theta} (y^n| x^n)$. Then $\int_{y^n} \max_{\theta} P_{M, \theta} (y^n|x^n) = \prod_{k \in [K]} \mathcal{R}(N_k, h_k)$, in which $\mathcal{R}(N_k, h_k)$ is the regret (denominator) of the NML distribution of the histogram model on the $k$th leaf node that contains $N_k$ data points and $h_k$ bins. }

\begin{proof}
    Consider the dataset $D=\{x^n, y^n\}$ and a CDTree with $K$ leaf nodes $M=\{S_1, ..., S_K\}$, each leaf $S_k$ equipped with a histogram model $H_k = (B_l, B_u, h_k)$, in which $h_k$ denotes the number of bins. Denote the histogram density estimator associated with each $H_k$ as $f_{H_k}(.)$, and in short $f_k(.)$, which contains a family of distributions parameterized by $\alpha^k$. Further, denote the subset of data points that end up in the $k$th leaf as $D_k = (x^{\{k\}}, y^{\{k\}}) := \{(x,y) \in D|x \in S_k\}$. Thus, 

    \begin{equation}
        \begin{split}
             \int_{y^n} \max_{\theta} P_{M, \theta} (y^n|x^n) & = \int_{y^n} \max_{\theta} \prod_{k \in [K]} f_k(y^{\{k\}}|x^{\{k\}}) \\
             & = \int_{y^n} \prod_{k \in [K]} \max_{\alpha^k}  f_k(y^{\{k\}}|x^{\{k\}}) \\
             % & = \int_{y^n} \prod_{k \in [K]} \max_{\alpha^k}  \prod_{j \in [h_k]} f_k(y^{\{k\}}_i|x^{\{k\}}_i) \\
              % & = \int_{y^n} \prod_{k \in [K]} \max_{\alpha^k}  \prod_{j \in [h_k]} (f_k(y^{\{k, j\}}|x^{\{k, j\}}))^{|D_{k,j}|} \\
               % & = \int_{y^n} \prod_{k \in [K]} \max_{\alpha^k}  \prod_{j \in [h_k]} f_k(y^{\{k\}}_i|x^{\{k\}}_i) \\
               \text{(Fubini's Theorem)} & = \int_{y^{\{1\}}}\int_{y^{\{2\}}}...\int_{y^{\{K\}}} \prod_{k \in [K]} \max_{\alpha^k}f_k(y^{\{k\}}|x^{\{k\}}) \\
               \text{(re-arranging)} & = \prod_{k \in [K]} \int_{y^{\{k\}}} \max_{\alpha^k} f_k(y^{\{k\}}|x^{\{k\}}) \\
               & = \prod_{k \in [K]} \mathcal{R}(N_k, h_k)
        \end{split}
    \end{equation}
    because the NML distribution for $D_k$ given the histogram $H_k$ is equal to~\citep{kontkanen2007mdl} 
    \begin{equation}
        P_{H_k}(y^{\{k\}}|x^{\{k\}}) = \frac{\max_{\alpha^k} f_{\alpha^k}(y^{\{k\}}|x^{\{k\}})} {\int_{y^{\{k\}}} \max_{\alpha^k} f_{\alpha^k}(y^{\{k\}}|x^{\{k\}}}.
    \end{equation}
    Further, it has been shown that the denominator (regret) of the NML distributions for histogram models is a function that only depends on $N_k$ and $h_k$~\citep{kontkanen2007mdl}. 
\end{proof}

\section{Algorithm Details for Optimizing Histogram Bins}\label{appendix:algs}

We further discuss the detail process of finding the number of histogram bins that minimizes the MDL score, for which the pseudo-code is provided in Algorithm~\ref{alg:hist}. 

While the search space for the number of bins, denoted as $h$, for the histogram can range from 1 to the number of data points, this is computationally expensive in practice. To mitigate this, we first narrow down the range of $h$ by fixing a step size $g$ and searching among $h \in \{1, g + 1, 2g + 1, \ldots\}$. The MDL-score will initially decrease as we increase the number of bins, since the the goodness-of-fit of the histogram increases. However, when the number of bins become too large, the MDL-based regularization terms will start to dominate, and hence the MDL-score starts to increase. 

Assuming the MDL-score starts to increase at $h = eg + 1$ (where $e$ is a positive integer), we can then narrow the range to $(e-2)g + 1 \leq h \leq eg + 1$, since the maximum so far is reached at $(e-1)g + 1$.

Finally, we iterate over all $h$ within this narrowed range, and we select the number of bins with the smallest MDL score. 

% While the search space of the number of bins, denoted as $h$, for the histogram can range from 1 to the number of data points, this in practice is too computationally expensive. Thus, we first try to narrow down the range of $h$ by fixing a step size $g$, and searching by among $h \in \{1, g + 1, 2g+1, ...\}$. The MDL-score will first decrease as we increase the number of bins, until a certain point the model complexity of the histogram becomes too large (i.e., when the NML-regret and the code length needed to encode the histogram start to dominate). Assume this happens at $h = eg + 1$ ($e$ is a positive integer), then we narrow down the range to $h \in \{(e-2)g + 1, ..., eg + 1\}$, since the maximum so far is reached at $(e-1)g + 1$. 

% Finally, we search the best number of bins with the smallest MDL-score among the range obtained in the previous step. 

\begin{algorithm}[ht]
\caption{Learn the MDL-optimal histogram} \label{alg:hist}
    \KwIn{Target values $y^{\{S\}}$ covered by a leaf node $S$, step size $g$}
    \KwOut{Histogram $H_S$ that minimizes the MDL-score}
    \BlankLine
    $h' \leftarrow 1$; 
    $best\_score \leftarrow +\infty$
    
    \While{True}{
        $mdl \leftarrow $ calculate the MDL-score given $h'$
        
        \eIf{$mdl < best\_score$}{
        $best\_score \leftarrow mdl$
        
        $h' \leftarrow h'  + g$
        }{
        break
        }
    }

    % \For{$i \in \max(1, h' - 2g) \,:\, h'$}{
    %     calculate the MDL-score given $i$ bins and find the number of bins $i$ that lead to the smallest MDL-score
    % }
    $bound\_low \leftarrow \max(1, h' - 2g)$ \tcp*{Since $h-2g$ may be smaller than 1}
    
    $bound\_high \leftarrow h'$
    
    $h^* \leftarrow$ the optimal $h$ within the range $bound\_low \leq h \leq bound\_high$ that minimizes the MDL-score
    
    \Return{$h^*$}
    
\end{algorithm}

\section{Experiment Details}\label{appendix:exp_setup}
\subsection{Experiment compute resources} \label{subsec:compute_resource}
The runtimes reported in Section~\ref{subsec:exp_runtime} for \emph{all} algorithms are recorded on the CPU machines with the AMD EPYC 7702 cores.

\subsection{Data preparation} We noticed that duplicated values in datasets may cause some of our competitor methods to fail. For instance, for those methods that adopt the Gaussian kernel, duplicated values can lead to the standard deviation estimated as 0 for some subsets of data points). Thus, we add a very small noise generated by $N(0, 10^{-3})$ to all data columns, \emph{which are used for all methods}.

\subsection{Parameters for CDTree} \label{sub_append:cdtree_parameters}
\paragraph{The parameter $C$.} 
We first discuss the parameter $C$ that controls the hierarchical structure for the search space of the splitting values, for which we set $C = 5$ in our experiments. 

The parameter $C$ can be used to express a prior belief on the candidate splitting values. For instance, when $C=5$, the code length needed for encoding the $5$ quantiles when $d=1$ are the same. Equivalently, the prior probabilities of these $5$ quantiles are the same. 

As these $5$ quantiles divide the values into $6$ subsets, essentially, we are saying that the first quantile among these 5 quantiles, i.e., the $1/6$-quantile, and the third quantile, i.e., the $1/2$-quantile ($3/6$-quantile), have the same prior probability. 

However, if we set $C=1$, there will be only 1 quantile when $d=1$, i.e., the $1/2$-quantile. In this case, the prior probability for the $1/2$-quantile will be different (larger) than the $1/6$-quantile.

Thus, the parameter $C$ can be used to express the prior belief in the splitting values. That is, one can ask herself that, is it equally likely, \emph{a priori}, to observe a node containing the splitting value equal to the $1/100$-quantile, and to observe a node containing the splitting value equal to the $1/2$-quantile (i.e., the median)?
% are a node containing the splitting value as the $1/100$-quantile and another node containing the splitting value as the $1/2$-quantile (i.e., the median) equally likely to be seen in the decision tree? 
If the answer is yes, $C$ might be set as $99$. Otherwise, a smaller $C$ should be considered. 

% Last, we comment that different choices on $C$ should have only very moderate influence on the CDTree learned from data. For instance, according to the discussion in Section~\ref{subsec:cl_model}, setting $C = 99$ makes the code length needed to encode the $1/100$-quantile equal to 
% \begin{equation}
%     L_\mathbf{N}(d=1) + \log_2(C=99) + d - 1, 
% \end{equation}
% which leads to $8.148$ bits. In contrast, setting $C=5$ requires $d = 8$ (as $2^{d-1} = 128$ in this case), which makes the code length needed to encode the $1/100$-quantile equal to 
% \begin{equation}
%     L_\mathbf{N}(d=8) + \log_2(C=5) + d - 1, 
% \end{equation}
% which leads to $16.08$ bits. 

\paragraph{The step size $g$.}
We set $g = 30$ in searching the number of histogram bins in Algorithm~\ref{alg:hist}. Note that if we set $g$ very large, we only need very few iterations to ``narrow" down the range for the number of bins for histograms; nevertheless, the resulting range will not be very narrow in this case. On the other hand, if we set $g$ very small, it may cost a lot of time to obtain the narrower range. In that sense, $g=30$ seems a rather balanced choice. 

\paragraph{The histogram boundary.}
Further, as discussed in Section~\ref{sec:disucssion}, we assume that the range of the target variable for each dataset is known. Thus, we set the global range of the histograms based on the range of the full dataset (before train/test split in the cross-validation) plus/minus a small constant, chosen as $10^{-3}$. 

\subsection{Details for competitor algorithms}
We use the implementation of CKDE, NKDE, LSCDE, NF, and MDN from from the Python package ``cde",\footnote{\url{https://github.com/freelunchtheorem/Conditional_Density_Estimation}}~\citep{rothfuss2019conditional} and the implementation of LinCDE and CADET from the original authors. 

\textbf{CADET.} No specific guidelines for regularization is provided in the original paper~\citep{cousins2019cadet}; further, in the author's original implementation, the standard regularization based on the tree size is not available. Instead, we can only choose the regularization mode in `constant', `AIC', and `BIC'. We pick the `BIC' as it behaves similarly to MDL asymptotically~\citep{grunwald2007minimum}. 

\textbf{CKDE.} The bandwidth is tuned by optimizing the cross-validation maximum likelihoods, which is implemented in `statsmodels'~\citep{seabold2010statsmodels}, the standard Python package for multivariate statistical analysis, and directly used in the `cde' package~\citep{rothfuss2019conditional}. For the three largest datasets (`support2', `navalprop', and `supercond'), we fail to obtain the cross-validation tuning results within 10 hours for a single fold, and hence we adopt the `normal reference' for the bandwidth selection.  

Note that it is not required to specify the range for searching the bandwidth, as the implementation is essentially based on the `optimize' function (for continuous optimization) in the well-known Python package `scipy'~\citep{2020SciPy-NMeth}. 

\textbf{NKDE.} The bandwidth and the $\epsilon$ (the parameter that controls the neighbor range) are tuned by optimizing the cross-validation maximum likelihoods, which is implemented in the Python package `cde'~\citep{rothfuss2019conditional}. Similar to CKDE, the search space for these two parameters are not required. For two very large datasets (`support2' and `supercond'), we fail to obtain the cross-validation tuning results within 10 hours for a single fold, and hence we adopt the `normal reference' for the bandwidth selection.  

\textbf{CART-k and CART-h.} In these methods, a CART regression tree is first trained, with the regularization hyperparameter for pruning tuned by cross-validation. After fixing the tree structure, a kernel density estimation model and a histogram model are fit to the subsets of data points on each leaf node for CART-k and CART-h, respectively. The bandwidth for the former is tuned by cross-validation, while the number of bins for the latter is picked to optimize the MDL score (given the fixed tree structure). Specifically, we use the CART implementation from the scikit-learn~\citep{scikit-learn} Python package. 

\textbf{LSCDE.} No tuning for bandwidth is discussed in the original paper, nor implemented. Thus, we stick to the default setting. 

\textbf{LinCDE.} The tree depth is set as $3$, which is a common setting for training boosted tree models.

\textbf{NF and MDN.} As suggested by ~\citet{rothfuss2019noisereg}, we use `noise regularization' for both NF and MDN. Specifically, we choose the standard deviation of added noise for both the features and the target as $0.01$. We further noticed that adding the standard dropout with dropout rate equal to $0.1$ gives more stable results. Other hyperparameters (e.g., the number of hidden layers) are kept as default. 

\section{More experiment results} \label{appendix:more_exp} 
\subsection{Standard deviation of negative log-likelihoods} \label{appendix:sd_nll}
In Table~\ref{table:sd_nll_appendix}, we report the negative log-likelihoods together with the standard deviations obtained by the five-fold cross-validation. In comparisons to other competitors, we observe no anomalously larger standard deviations for our method. 

\input{nll_table_sd.tex}

\subsection{Robustness against irrelevant features} \label{appendix:scala}
\begin{figure}[ht]
    \centering
    \includegraphics[width=\textwidth]{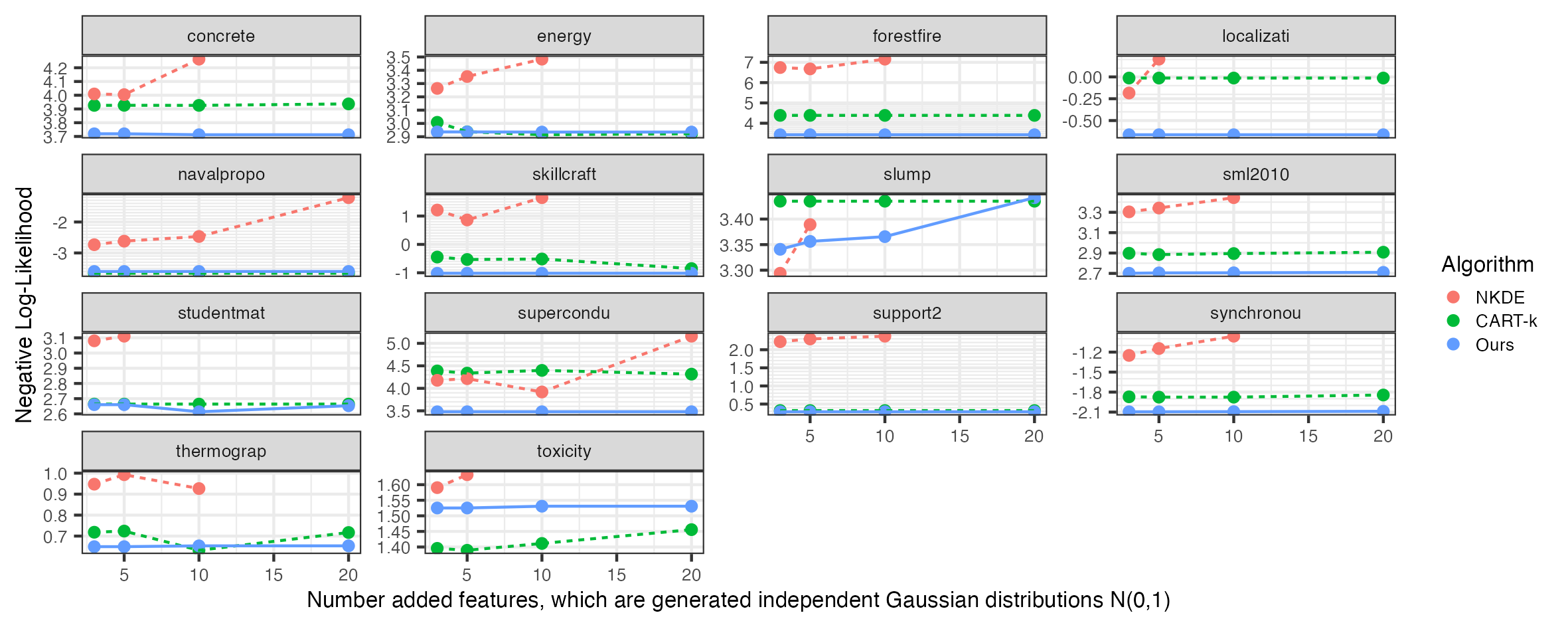}
    \caption{Negative-log-likelihoods with different number of added features, which are generated by independent Gaussian distributions. The results of CDTree, shown in blue solid lines, are extremely stable on all datasets expect for the very small `slump' dataset. }
    \label{fig:scala_more}
\end{figure}

In Figure~\ref{fig:scala_more}, we show the results that demonstrate the stability of the negative log-likelihoods when independent gaussian features are added to the 14 UCI datasets, in which we observe similar results as in Figure~\ref{fig:scala}, i.e., our proposed method CDTree is extremely stable across all datasets except for the very small dataset `slump'. 

\section{Medical costs visualizations} \label{appendix:medical}
\begin{figure}[!h]
    \centering
    \includegraphics[width=\textwidth]{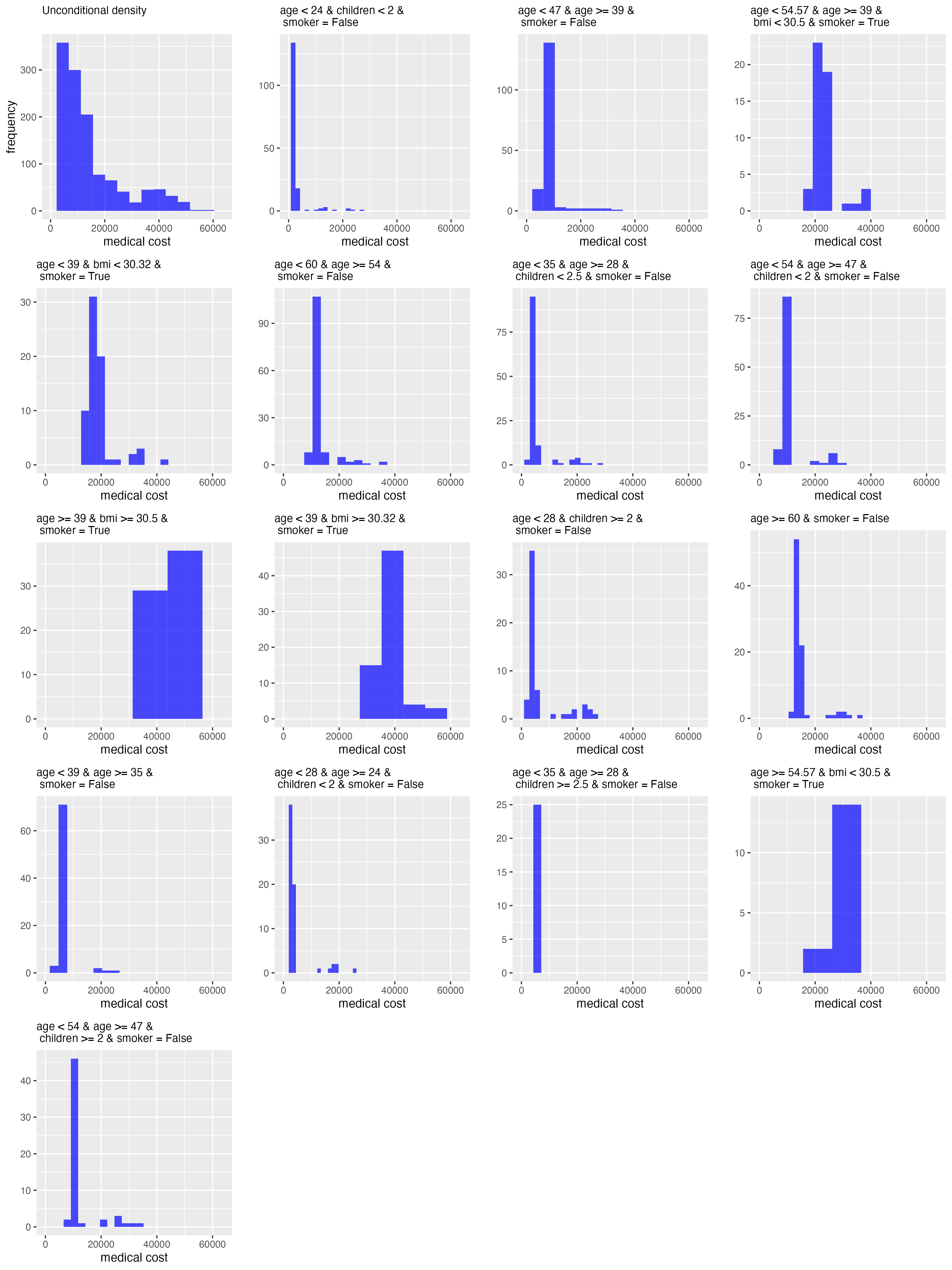}
    \caption{All leaves from the CDTree that describes the conditional density of medical costs, conditioned on demographic features. Root-to-leaf conditions are presented after removing logically redundant conditions.}
    \label{fig:meidcal_more}
\end{figure}

We show in Figure~\ref{fig:meidcal_more} the full results of applying CDTree to the medical expenses dataset used in Section~\ref{sec:intro}. As a sanity check for the goodness-of-fit, we report the average cross-validation negative log-likelihood (on test sets) for CDTree is $9.03$ (with the standard deviation $0.13$), which is slightly better than that of the black-box boosted tree model LinCDE, with the average negative log-likelihood $9.55$ (standard deviation $0.09$). Each Histogram corresponds to a single leaf that describes a meaningful subgroup of patients, in which we observe `age', `BMI', and `smoker' are the most distinguishing features to characterize each subgroup. Last, although the dataset contains 9 features and only 1338 samples, the illustrations show that CDTree does not require a very large dataset to reveal meaningful and comprehensible information. 

\newpage
\section*{NeurIPS Paper Checklist}

%%% BEGIN INSTRUCTIONS %%%
The checklist is designed to encourage best practices for responsible machine learning research, addressing issues of reproducibility, transparency, research ethics, and societal impact. Do not remove the checklist: {\bf The papers not including the checklist will be desk rejected.} The checklist should follow the references and follow the (optional) supplemental material.  The checklist does NOT count towards the page
limit. 

Please read the checklist guidelines carefully for information on how to answer these questions. For each question in the checklist:
\begin{itemize}
    \item You should answer \answerYes{}, \answerNo{}, or \answerNA{}.
    \item \answerNA{} means either that the question is Not Applicable for that particular paper or the relevant information is Not Available.
    \item Please provide a short (1–2 sentence) justification right after your answer (even for NA). 
   % \item {\bf The papers not including the checklist will be desk rejected.}
\end{itemize}

{\bf The checklist answers are an integral part of your paper submission.} They are visible to the reviewers, area chairs, senior area chairs, and ethics reviewers. You will be asked to also include it (after eventual revisions) with the final version of your paper, and its final version will be published with the paper.

The reviewers of your paper will be asked to use the checklist as one of the factors in their evaluation. While "\answerYes{}" is generally preferable to "\answerNo{}", it is perfectly acceptable to answer "\answerNo{}" provided a proper justification is given (e.g., "error bars are not reported because it would be too computationally expensive" or "we were unable to find the license for the dataset we used"). In general, answering "\answerNo{}" or "\answerNA{}" is not grounds for rejection. While the questions are phrased in a binary way, we acknowledge that the true answer is often more nuanced, so please just use your best judgment and write a justification to elaborate. All supporting evidence can appear either in the main paper or the supplemental material, provided in appendix. If you answer \answerYes{} to a question, in the justification please point to the section(s) where related material for the question can be found.

%%% END INSTRUCTIONS %%%

\begin{enumerate}

\item {\bf Claims}
    \item[] Question: Do the main claims made in the abstract and introduction accurately reflect the paper's contributions and scope?
    \item[] Answer: \answerYes{}%\answerTODO{} % Replace by \answerYes{}, \answerNo{}, or \answerNA{}.
    \item[] Justification: Our mains claims and the paper's contributions are both 1) to introduce a tree-based CDE method, and 2) we show that our proposed method is both more accurate (as measured by the log-loss), has smaller tree sizes, and is robust against irrelevant features. %\justificationTODO{}

\item {\bf Limitations}
    \item[] Question: Does the paper discuss the limitations of the work performed by the authors?
    \item[] Answer: \answerYes{}%\answerTODO{} % Replace by \answerYes{}, \answerNo{}, or \answerNA{}.
    \item[] Justification: We have a separate `limitation' subsection within the Discussion Section in Section~\ref{sec:disucssion}.%\justificationTODO{}

\item {\bf Theory Assumptions and Proofs}
    \item[] Question: For each theoretical result, does the paper provide the full set of assumptions and a complete (and correct) proof?
    \item[] Answer: \answerYes{}%\answerTODO{} % Replace by \answerYes{}, \answerNo{}, or \answerNA{}.
    \item[] Justification: our proof has no additional assumptions.  %\justificationTODO{}

    \item {\bf Experimental Result Reproducibility}
    \item[] Question: Does the paper fully disclose all the information needed to reproduce the main experimental results of the paper to the extent that it affects the main claims and/or conclusions of the paper (regardless of whether the code and data are provided or not)?
    \item[] Answer: \answerYes{}%\answerTODO{} % Replace by \answerYes{}, \answerNo{}, or \answerNA{}.
    \item[] Justification: Full descriptions are provided both in the main context and in the appendix. Also, the source code is provided. %\justificationTODO{}

\item {\bf Open access to data and code}
    \item[] Question: Does the paper provide open access to the data and code, with sufficient instructions to faithfully reproduce the main experimental results, as described in supplemental material?
    \item[] Answer: \answerYes{}%\answerTODO{} % Replace by \answerYes{}, \answerNo{}, or \answerNA{}.
    \item[] Justification: The source code is provided. The datasets are all available on the website of the UCI repository. %\justificationTODO{}

\item {\bf Experimental Setting/Details}
    \item[] Question: Does the paper specify all the training and test details (e.g., data splits, hyperparameters, how they were chosen, type of optimizer, etc.) necessary to understand the results?
    \item[] Answer: \answerYes{}%\answerTODO{} % Replace by \answerYes{}, \answerNo{}, or \answerNA{}.
    \item[] Justification:  Full descriptions are provided both in the main context and in the appendix.
    %\justificationTODO{}

\item {\bf Experiment Statistical Significance}
    \item[] Question: Does the paper report error bars suitably and correctly defined or other appropriate information about the statistical significance of the experiments?
    \item[] Answer: \answerYes{}%\answerTODO{} % Replace by \answerYes{}, \answerNo{}, or \answerNA{}.
    \item[] Justification: we report the standard deviation of the negative log-likelihoods, obtained by five-fold cross-validation, in Appendix~\ref{appendix:sd_nll}. %Justification: Our proposed algorithm is not a stochastic algorithm. Although some randomness is introduced as we report our results on the test sets based on the five-fold cross-validation, we believe the influence is minimal and won't affect our conclusion. %\justificationTODO{}

\item {\bf Experiments Compute Resources}
    \item[] Question: For each experiment, does the paper provide sufficient information on the computer resources (type of compute workers, memory, time of execution) needed to reproduce the experiments?
    \item[] Answer: \answerYes{} %\answerTODO{} % Replace by \answerYes{}, \answerNo{}, or \answerNA{}.
    \item[] Justification: We report the compute resources in Appendix~\ref{subsec:compute_resource}. %\justificationTODO{}
    
\item {\bf Code Of Ethics}
    \item[] Question: Does the research conducted in the paper conform, in every respect, with the NeurIPS Code of Ethics \url{https://neurips.cc/public/EthicsGuidelines}?
    \item[] Answer: \answerYes{}%\answerTODO{} % Replace by \answerYes{}, \answerNo{}, or \answerNA{}.
    \item[] Justification: \answerNA%\justificationTODO{}

\item {\bf Broader Impacts}
    \item[] Question: Does the paper discuss both potential positive societal impacts and negative societal impacts of the work performed?
    \item[] Answer: \answerYes{}%\answerTODO{} % Replace by \answerYes{}, \answerNo{}, or \answerNA{}.
    \item[] Justification: The paper discusses the positive societal impacts about 1) interpretable CDE methods have advantages in critical domains, 2) CDTree is more interpretable than kernel-based methods because it is also interpretable to the general public without Statistics knowledge. As far as we are concerned, our method does not have the risk of being misused in a malicious way. 
    
    
\item {\bf Safeguards}
    \item[] Question: Does the paper describe safeguards that have been put in place for responsible release of data or models that have a high risk for misuse (e.g., pretrained language models, image generators, or scraped datasets)?
    \item[] Answer: \answerNA%\answerTODO{} % Replace by \answerYes{}, \answerNo{}, or \answerNA{}.
    \item[] Justification: \answerNA %\justificationTODO{}

\item {\bf Licenses for existing assets}
    \item[] Question: Are the creators or original owners of assets (e.g., code, data, models), used in the paper, properly credited and are the license and terms of use explicitly mentioned and properly respected?
    \item[] Answer: \answerYes{}
    % \answerTODO{} % Replace by \answerYes{}, \answerNo{}, or \answerNA{}.
    \item[] Justification: All datasets and code are free to use for research purposes according to the licenses. 

\item {\bf New Assets}
    \item[] Question: Are new assets introduced in the paper well documented and is the documentation provided alongside the assets?
    \item[] Answer: \answerYes{} % Replace by \answerYes{}, \answerNo{}, or \answerNA{}.
    \item[] Justification: Source code is provided, together with the details of how to get the code running. 

\item {\bf Crowdsourcing and Research with Human Subjects}
    \item[] Question: For crowdsourcing experiments and research with human subjects, does the paper include the full text of instructions given to participants and screenshots, if applicable, as well as details about compensation (if any)? 
    \item[] Answer: \answerNA%\answerTODO{} % Replace by \answerYes{}, \answerNo{}, or \answerNA{}.
    \item[] Justification: \answerNA %\justificationTODO{}

\item {\bf Institutional Review Board (IRB) Approvals or Equivalent for Research with Human Subjects}
    \item[] Question: Does the paper describe potential risks incurred by study participants, whether such risks were disclosed to the subjects, and whether Institutional Review Board (IRB) approvals (or an equivalent approval/review based on the requirements of your country or institution) were obtained?
    \item[] Answer: \answerNA %\answerTODO{} % Replace by \answerYes{}, \answerNo{}, or \answerNA{}.
    \item[] Justification: \answerNA %\justificationTODO{}

\end{enumerate}

\end{document}

%% file: data_info_tab.tex
% \begin{table}[ht]
% \centering
% \caption{Datasets from the UCI repository, with their number of rows and columns} \label{table:data_info}
% \begin{tabular}{lrr|lrr}
%   \hline
% dataset & nrow & ncol & dataset & nrow & ncol \\ 
%   \hline
% energy & 9568 &   5 & navalpropo & 11934 &  17 \\ 
%   synchronou & 557 &   5 & skillcraft & 3338 &  19 \\ 
%   localizati & 107 &   6 & sml2010 & 2764 &  21 \\ 
%   toxicity & 908 &   7 & thermograp & 1003 &  26 \\ 
%   concrete & 1030 &   9 & support2 & 9105 &  27 \\ 
%   slump & 103 &  10 & studentmat & 395 &  44 \\ 
%   forestfire & 517 &  13 & supercondu & 21263 &  82 \\ 
%    \hline
% \end{tabular}
% % \vspace*{+3mm}
% % 
% \end{table}
\begin{table}[ht]
\centering
\caption{Datasets from the UCI repository, with the numbers of rows and columns} \label{table:data_info}
\resizebox{\textwidth}{!}{
\begin{tabular}{lrrrrrrrr}
  \hline
dataset & energy & synchronou & localizati & toxicity & concrete & slump & forestfire \\ 
\# rows & 9568 & 557 & 107 & 908 & 1030 & 103 & 517 \\ 
\# cols & 5 & 5 & 6 & 7 & 9 & 10 & 13 \\ 
  \hline
dataset & navalpropo & skillcraft & sml2010 & thermograp & support2 & studentmat & supercondu \\ 
\# rows & 11934 & 3338 & 2764 & 1003 & 9105 & 395 & 21263 \\ 
\# cols & 17 & 19 & 21 & 26 & 27 & 44 & 82 \\ 
   \hline
\end{tabular}
}
% \vspace*{+3mm}
% 
\end{table}

%% file: exp1_all_tab.tex
\begin{table}[t]
\centering
\caption{Negative log-likelihoods (smaller is better) on test sets. The best results among interpretable methods are shown in \textbf{bold}, and the best results among all interpretable and black-box models are marked by the \underline{underlines}. The datasets are ordered by their numbers of columns (ascending).} \label{table:cde_loglikelihood}
\begin{tabular}{lrrrrrrr|rrr}
  \hline
    & \multicolumn{7}{c|}{Interpretable models} &  \multicolumn{3}{c}{Black-box models} \\ 
\hline
\scriptsize{Datasets} & \scriptsize{CADET} & \scriptsize{CART-h} & \scriptsize{CART-k} & \scriptsize{CKDE} & \scriptsize{LSCDE} & \scriptsize{NKDE} & \emph{\scriptsize{Ours}} & \scriptsize{LinCDE} & \scriptsize{MDN} & \scriptsize{NF} \\ 
  \hline
energy & 3.55 & 3.09 & 3.06 & \textbf{\underline{2.47}} & 3.38 & 3 & 2.93 & 2.93 & 2.78 & 2.86 \\ 
  synchrono & -2.93 & -1.63 & -1.86 & \textbf{\underline{-3.59}} & -1.25 & -1.57 & -2.11 & -1.85 & -2.94 & -2.64 \\ 
  localizat & -0.23 & -0.55 & -0.01 & -0.26 & -0.61 & -0.28 & \textbf{-0.66} & \underline{-0.95} & -0.68 & -0.43 \\ 
  toxicity & 1.8 & 1.5 & 1.38 & \textbf{1.32} & 1.34 & 1.55 & 1.53 & 1.29 & 1.24 & \underline{1.23} \\ 
  concrete & 4.17 & 3.75 & 3.93 & \textbf{3.32} & 3.66 & 3.91 & 3.72 & 3.47 & \underline{2.97} & 3.18 \\ 
  slump & 3.42 & 3.55 & 3.43 & \textbf{2.35} & 2.91 & 3.08 & 3.34 & 2.98 & \underline{2.23} & 2.39 \\ 
  forestfir & 134 & 3.96 & 4.39 & 4.85 & 4.68 & 5.55 & \textbf{3.43} & 4.35 & 3.26 & \underline{3.23} \\ 
  navalprop & -3.53 & -3.3 & \textbf{-3.66} & -2.8 & -2.88 & -3.19 & -3.6 & -3.36 & \underline{-4.12} & -3.75 \\ 
  skillcraf & 94.4 & 0.46 & -0.42 & 1.54 & 1.61 & 1.56 & \textbf{\underline{-1.02}} & 1.26 & 0.35 & 1.11 \\ 
  sml2010 & 6.52 & 2.85 & 2.89 & \textbf{\underline{1.61}} & 3.14 & 3.12 & 2.7 & 2.97 & 2.15 & 2.61 \\ 
  thermogra & 2.21 & 0.66 & 0.72 & 0.66 & 0.94 & 0.94 & \textbf{0.64} & 0.59 & 0.56 & \underline{0.52} \\ 
  support2 & 97.3 & 0.51 & 0.32 & 2.09 & 2.46 & 2.13 & \textbf{\underline{0.29}} & 1.48 & 1.53 & 1.24 \\ 
  studentma & 3.83 & \textbf{2.65} & 2.66 & 2.89 & 4.19 & 3.11 & 2.66 & \underline{2.59} & 3.85 & 3.54 \\ 
  supercond & 9.6 & 3.84 & 4.36 & 4.55 & 4.17 & 4.19 & \textbf{3.48} & 3.87 & \underline{3.33} & 3.5 \\ 
  \hline
  rank (all) & 8.79 & 5.68 & 6.04 & 5.11 & 7.46 & 7.68 & 4 & 4.46 & \textbf{2.57} & 3.21 \\ 
   \hline
   rank (intp.) & 6.07 & 3.43 & 3.86 & 3.14 & 4.57 & 4.86 & \textbf{2.07} & --- & --- & --- \\
  \hline
\end{tabular}
% \vspace*{+3mm}

\end{table}

%% file: nll_table_sd.tex
\begin{table}[ht]
\caption{Negative log-likelihoods (smaller is better) on test sets, together with the standard deviation, obtained using five-fold cross-validation.} \label{table:sd_nll_appendix}
\centering
\resizebox{\textwidth}{!}{
\begin{tabular}{lrrrrrrr|rrr}
  \hline
    & \multicolumn{7}{c|}{Interpretable models} &  \multicolumn{3}{c}{Black-box models} \\ 
  \hline
dataset & CADET & CART-h & CART-k & CKDE & LSCDE & NKDE & Ours & LinCDE & MDN & NF \\ 
  \hline
energy & 3.55 (0.22) & 3.09 (0.01) & 3.06 (0.17) & 2.47 (0.02) & 3.38 (0.01) & 3 (0.05) & 2.93 (0.01) & 2.93 (0.01) & 2.78 (0.02) & 2.86 (0.05) \\ 
  synchronou & -2.93 (0.24) & -1.63 (0.07) & -1.86 (0.1) & -3.59 (0.09) & -1.25 (0.02) & -1.57 (0.04) & -2.11 (0.02) & -1.85 (0.03) & -2.94 (0.05) & -2.64 (0.15) \\ 
  localizati & -0.23 (0.86) & -0.55 (0.29) & -0.01 (1.63) & -0.26 (1.79) & -0.61 (0.76) & -0.28 (0.98) & -0.66 (0.27) & -0.95 (0.11) & -0.68 (0.46) & -0.43 (0.95) \\ 
  toxicity & 1.8 (0.19) & 1.5 (0.09) & 1.38 (0.09) & 1.32 (0.05) & 1.34 (0.02) & 1.55 (0.08) & 1.53 (0.09) & 1.29 (0.06) & 1.24 (0.08) & 1.23 (0.07) \\ 
  concrete & 4.17 (0.47) & 3.75 (0.03) & 3.93 (0.57) & 3.32 (0.06) & 3.66 (0.03) & 3.91 (0.07) & 3.72 (0.06) & 3.47 (0.03) & 2.97 (0.06) & 3.18 (0.13) \\ 
  slump & 3.42 (0.71) & 3.55 (0.19) & 3.43 (0.17) & 2.35 (0.2) & 2.91 (0.2) & 3.08 (0.25) & 3.34 (0.22) & 2.98 (0.15) & 2.23 (0.33) & 2.39 (0.18) \\ 
  forestfire & 133.95 (106.29) & 3.96 (0.12) & 4.39 (1.35) & 4.85 (0.72) & 4.68 (0.14) & 5.55 (0.12) & 3.43 (0.19) & 4.35 (0.24) & 3.26 (0.48) & 3.23 (0.74) \\ 
  navalpropo & -3.53 (0.12) & -3.3 (0.04) & -3.66 (0.07) & -2.8 (0) & -2.88 (0.01) & -3.19 (0.22) & -3.6 (0.06) & -3.36 (0) & -4.12 (0.04) & -3.75 (0.2) \\ 
  skillcraft & 94.44 (46.08) & 0.46 (0.7) & -0.42 (0.53) & 1.54 (0.01) & 1.61 (0.02) & 1.56 (0.01) & -1.02 (0.03) & 1.26 (0.01) & 0.35 (1.26) & 1.11 (0.27) \\ 
  sml2010 & 6.52 (2.65) & 2.85 (0.05) & 2.89 (0.43) & 1.61 (0.17) & 3.14 (0.01) & 3.12 (0.04) & 2.7 (0.06) & 2.97 (0.02) & 2.15 (0.02) & 2.61 (0.17) \\ 
  thermograp & 2.21 (0.62) & 0.66 (0.05) & 0.72 (0.35) & 0.66 (0.06) & 0.94 (0.05) & 0.94 (0.05) & 0.64 (0.02) & 0.59 (0.06) & 0.56 (0.19) & 0.52 (0.03) \\ 
  support2 & 97.33 (81.6) & 0.51 (0.04) & 0.32 (0.06) & 2.09 (0.01) & 2.46 (0.08) & 2.13 (0.02) & 0.29 (0.04) & 1.48 (0.07) & 1.53 (0.86) & 1.24 (0.22) \\ 
  studentmat & 3.83 (0.39) & 2.65 (0.05) & 2.66 (0.05) & 2.89 (0.06) & 4.19 (0.41) & 3.11 (0.11) & 2.66 (0.07) & 2.59 (0.05) & 3.85 (0.34) & 3.54 (0.36) \\ 
  supercondu & 9.6 (2.73) & 3.84 (0.01) & 4.36 (0.14) & 4.55 (0) & 4.17 (0.02) & 4.19 (0.15) & 3.48 (0.02) & 3.87 (0.02) & 3.33 (0.03) & 3.5 (0.04) \\ 
   \hline
\end{tabular}
}
\end{table}